\PassOptionsToPackage{table}{xcolor}
\documentclass[sigconf]{acmart}

\usepackage{subcaption}
\usepackage{amsmath}
\usepackage{mathtools}
\usepackage{amsthm}
\usepackage[capitalize,noabbrev]{cleveref}
\usepackage{algorithm}
\usepackage{algpseudocode}
\usepackage{tabularx}
\usepackage{array}

\newcolumntype{Y}{>{\centering\arraybackslash}X}

\AtEndPreamble{%
  \theoremstyle{acmdefinition}

}

\AtBeginDocument{%
  }

\copyrightyear{2026}
\acmYear{2026}
\setcopyright{cc}
\setcctype{by}
\acmConference[MM '26]{Proceedings of the 34th ACM International Conference on Multimedia}{November 10--14, 2026}{Rio de Janeiro, Brazil}
\acmBooktitle{Proceedings of the 34th ACM International Conference on Multimedia (MM '26), November 10--14, 2026, Rio de Janeiro, Brazil}
\acmDOI{10.1145/3767308.3835356}
\acmISBN{979-8-4007-2213-4/2026/11}
\title{When Classes Evolve: A Benchmark and Framework for Stage-Aware Class-Incremental Learning}

\author{Zheng Zhang}
\authornote{Co-first authors contributed equally.}
\affiliation{%
  \institution{Hefei Institutes of Physical Science, Chinese Academy of Sciences}
  \city{Hefei}
  \state{Anhui}
  \country{China}}
\affiliation{%
  \institution{Dalian University of Technology}
  \city{Dalian}
  \state{Liaoning}
  \country{China}}
\email{ericzhengz@mail.dlut.edu.cn}

\author{Tao Hu}
\authornotemark[1]
\affiliation{%
  \institution{Hefei Institutes of Physical Science, Chinese Academy of Sciences}
  \city{Hefei}
  \state{Anhui}
  \country{China}}
\affiliation{%
  \institution{University of Science and Technology of China}
  \city{Hefei}
  \state{Anhui}
  \country{China}}
\email{ht\_simon@mail.ustc.edu.cn}

\author{Xueheng Li}
\authornotemark[1]
\affiliation{%
  \institution{Hefei Institutes of Physical Science, Chinese Academy of Sciences}
  \city{Hefei}
  \state{Anhui}
  \country{China}}
\affiliation{%
  \institution{University of Science and Technology of China}
  \city{Hefei}
  \state{Anhui}
  \country{China}}
\email{lixueheng@mail.ustc.edu.cn}

\author{Yang Wang}
\affiliation{%
  \institution{Hefei Institutes of Physical Science, Chinese Academy of Sciences}
  \city{Hefei}
  \state{Anhui}
  \country{China}}
\affiliation{%
  \institution{University of Science and Technology of China}
  \city{Hefei}
  \state{Anhui}
  \country{China}}
\email{yangwang00@mail.ustc.edu.cn}

\author{Rui Li}
\affiliation{%
  \institution{Hefei Institutes of Physical Science, Chinese Academy of Sciences}
  \city{Hefei}
  \state{Anhui}
  \country{China}}
\email{lirui@iim.ac.cn}

\author{Jie Zhang}
\authornote{Corresponding authors.}
\affiliation{%
  \institution{Hefei Institutes of Physical Science, Chinese Academy of Sciences}
  \city{Hefei}
  \state{Anhui}
  \country{China}}
\affiliation{%
  \institution{Zhongke Hefei Institute of Technology Innovation Engineering}
  \city{Hefei}
  \state{Anhui}
  \country{China}}
\email{zhangjie@iim.ac.cn}

\author{Chengjun Xie}
\authornotemark[2]
\affiliation{%
  \institution{Hefei Institutes of Physical Science, Chinese Academy of Sciences}
  \city{Hefei}
  \state{Anhui}
  \country{China}}
\affiliation{%
  \institution{Zhongke Hefei Institute of Technology Innovation Engineering}
  \city{Hefei}
  \state{Anhui}
  \country{China}}
\email{cjxie@iim.ac.cn}

\renewcommand{\shortauthors}{Zheng Zhang et al.}

\begin{document}

\begin{abstract}
Class-Incremental Learning (CIL) aims to sequentially learn new classes while mitigating catastrophic forgetting of previously learned knowledge. Conventional CIL approaches implicitly assume that classes are morphologically static, focusing primarily on preserving previously learned representations as new classes are introduced. In practice, however, instances of the same semantic class may undergo substantial morphological evolution, such as a larva turning into a butterfly. Consequently, a model must both discriminate between classes and adapt to evolving appearances within a single class. To systematically address this challenge, we formalize Stage-Aware CIL (Stage-CIL), a paradigm in which each class is learned progressively through distinct morphological stages. We further introduce Stage-Bench, a 10-domain, two-stage benchmark and protocol for evaluating both inter-class forgetting and stage-level degradation within classes. Finally, we propose STAGE, an evolution-aware reference baseline that disentangles semantic identity from evolution dynamics through a fixed-size memory pool, enabling stage-aware prediction of later morphological forms from earlier representations. Extensive experiments show that conventional CIL reductions and existing continual-learning baselines remain insufficient under Stage-CIL, while STAGE consistently outperforms strong competitors, demonstrating the promise of explicit evolution-aware modeling for this new setting.
\end{abstract}


\maketitle

\section{Introduction}

Intelligent systems deployed in the real world must learn from non-stationary data streams \cite{shaheen2022continual, wang2024comprehensive}. Class-Incremental Learning (CIL) \cite{rebuffi2017icarl, de2021continual} addresses this requirement by sequentially introducing new classes while retaining previously acquired knowledge, yet remains fundamentally constrained by catastrophic forgetting \cite{french1999catastrophic, mccloskey1989catastrophic}. Contemporary solutions, including replay-based strategies \cite{choi2024dslr, jiang2025dupt} and PTM-driven modular tuning \cite{wang2022learning, sun2025mos, zhou2025learning,zhang2026btspcam}, predominantly focus on mitigating inter-class interference to preserve previously learned categories. This principle is effective for inter-class discrimination, ensuring that learning ``fish'' does not erase ``butterflies'', but implicitly assumes that each class remains morphologically static over time\cite{hu2024causality, hu2026pestclip}.
\begin{figure}[t]
\centering
\includegraphics[width=0.9\linewidth]{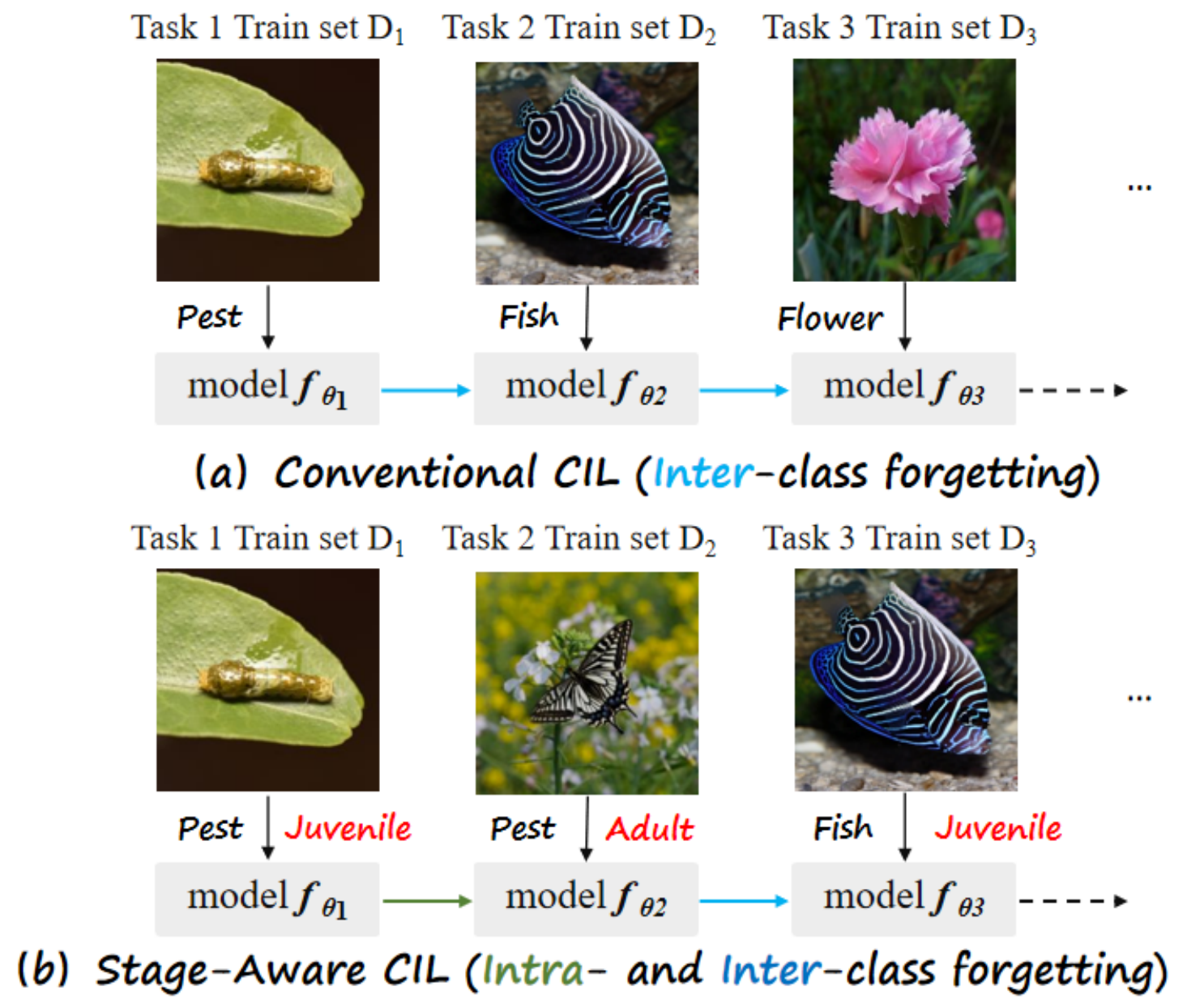}
\caption{Conventional CIL focuses on inter-class forgetting across distinct categories. Stage-CIL introduces temporally ordered morphological stages within each class, giving rise to the new problem of intra-class forgetting while retaining the original inter-class challenge.}
\label{FIG:1}
\end{figure}

In many natural and man-made scenarios, the same semantic class may undergo substantial stage-specific transformation. A larva becomes a butterfly; a new car gradually accumulates damage. In such cases, the core challenge is not merely adapting to visual variation, but explicitly preserving identity consistency across markedly different stage-specific forms\cite{li2026pestthinkerlearningthinkreason, li2026pestvlnetenablingmultimodalpest}. Naively applying conventional CIL protocols to this scenario creates a fundamental dilemma: grouping all stages under a single static class label flattens the evolutionary structure and induces severe intra-class feature interference, whereas splitting them into separate pseudo-classes severs their underlying identity link. As a result, models struggle to preserve a coherent semantic identity across stages, giving rise to a new form of intra-class forgetting (Fig.~\ref{FIG:1}). Standard domain-shift protocols do not explicitly evaluate this class-conditioned identity preservation within an expanding label space, leaving this structured intra-class evolution insufficiently isolated by current CIL and DIL benchmarks.

We therefore formalize Stage-Aware Class-Incremental Learning (Stage-CIL), a new setting in which the same semantic class is encountered through distinct morphological stages during class-incremental learning. To support rigorous study, we construct Stage-Bench, the first multi-domain benchmark tailored to Stage-CIL. Spanning 10 diverse domains, it provides standardized protocols, evaluation splits, and two dedicated diagnostic metrics: Inter- and Intra-Class Forgetting.

To further validate that Stage-CIL is learnable, we introduce STAGE, an evolution-aware reference baseline. STAGE disentangles stable semantic identity from stage dynamics through a shared set of canonical patterns and semantic anchors, enabling cross-stage prediction across stage-specific forms. In this way, STAGE serves as a strong proof-of-concept empirical baseline for this new setting.

Our contributions are threefold:

\begin{itemize}
\item We introduce Stage-CIL, a new continual-learning setting and principled protocol that preserves identity across intra-class stage evolution.
\item We establish Stage-Bench, the first benchmark for Stage-CIL, with standardized evaluation tools and diagnostic metrics tailored to this new setting.
\item We develop STAGE as a baseline, showing that explicit modeling of stage transformation is a promising direction for addressing intra-class forgetting in Stage-CIL.
\end{itemize}
\section{Related Work}
\subsection{Class-Incremental Learning}

Class-Incremental Learning (CIL) studies the problem of learning a sequence of class sets while retaining knowledge of previously observed classes~\cite{rebuffi2017icarl}. Existing methods primarily address catastrophic forgetting through several representative paradigms, including distillation-based regularization~\cite{LiHoiem2018, rebuffi2017icarl}, rehearsal with exemplar replay~\cite{ChaudhryRWalk18}, parameter regularization such as EWC~\cite{KirkpatrickSI17}, and model expansion with task-specific components~\cite{rusu2016progressive}.

With the increasing adoption of Pre-trained Models (PTMs), PTM-based CIL has become a central research direction~\cite{zhou2024continual}. These methods typically freeze the backbone and focus on adapting lightweight modules. L2P~\cite{wang2022learning} retrieves instance-specific prompts from a prompt pool, and DualPrompt~\cite{wang2022dualprompt} separates task-invariant and task-specific prompts. CODA-Prompt~\cite{smith2023coda} composes decomposed prompts using attention mechanisms. Adapter-based designs such as MOS~\cite{sun2025mos} and TUNA~\cite{wang2025integrating} introduce lightweight task-specific or shared adaptation modules. Prototype-driven approaches, including SimpleCIL~\cite{zhou2025revisiting}, APER~\cite{zhou2024aper}, and RanPAC~\cite{McDonnell2023RanPAC}, leverage prototypical classifiers or random projections on top of frozen features, often achieving strong performance with minimal parameter updates. Overall, existing CIL methods primarily focus on retaining previously learned categories under an expanding label space.




\subsection{Domain-Incremental Learning}

Domain-Incremental Learning (DIL) studies scenarios in which the label space remains fixed while the input distribution changes across domains~\cite{vandeVenTolias22}. The goal is to acquire new domain knowledge while maintaining performance on previously learned domains, often without relying on explicit task identifiers at test time. S-Prompts~\cite{Wang2022SsPrompts} learns a pool of domain-specific prompts and retrieves them by similarity search, while DCE~\cite{Li2025DCE} introduces frequency-aware expert networks with a dynamic selector to balance intra- and inter-domain adaptation. Prototype- or projection-based methods such as SimpleCIL~\cite{zhou2025revisiting} and RanPAC~\cite{McDonnell2023RanPAC} have also shown strong cross-domain performance with minimal parameter updates. Overall, standard DIL primarily focuses on preserving recognition under domain-wise distribution shifts with a fixed label space.

\section{Benchmark Settings}

\subsection{Formulation}

In standard Class-Incremental Learning, a model is trained on a sequence of tasks $\mathcal{T}=\{T_1,T_2,\dots,T_{N_{\mathrm{task}}}\}$. Each task $T_t$ provides a dataset $D_t=\{(x_i,y_i)\}_{i=1}^{n_t}$, where $x_i$ is an input sample and $y_i\in\mathcal{C}_t$ is its class label. The class sets are disjoint across tasks, i.e., $\mathcal{C}_t\cap\mathcal{C}_{t'}=\emptyset$ for $t\neq t'$, and the cumulative label space up to task $t$ is denoted by $\mathcal{C}_{1:t}=\bigcup_{j=1}^{t}\mathcal{C}_j$. During training on $T_t$, the model only has access to $D_t$. The objective is maintaining strong performance on all classes in $\mathcal{C}_{1:t}$ while mitigating catastrophic forgetting of earlier knowledge.

Stage-Aware Class-Incremental Learning extends this setting by associating each sample with a stage annotation. Formally, each task $T_t$ provides a dataset $D_t=\{(x_i,y_i,s_i)\}_{i=1}^{n_t}$, where $x_i$ is the input, $y_i\in\mathcal{C}_t$ is the class label, and $s_i\in\mathcal{S}_{y_i}$ denotes the stage label associated with class $y_i$. Here, $\mathcal{S}_{c}$ is the finite set of admissible stages for class $c$. The prediction target remains the class label $y_i$, while the stage label $s_i$ specifies the evolution state under which class $y_i$ is observed. Treating $(y_i,s_i)$ as independent labels would collapse the problem into stage-specific fine-grained classification, thereby removing the requirement of preserving semantic identity across stages. For notational convenience, when all classes share a common stage index set, we write $\mathcal{S}_{c}=\mathcal{S}$ for all $c$.

For any task $T_t$, we denote by
\[
D_t^{(c,s)}=\{(x_i,y_i,s_i)\in D_t \mid y_i=c,\ s_i=s\}
\]
the subset of samples in $D_t$ associated with class $c$ and stage $s$. Accordingly,
\[
D_t=\bigcup_{c\in\mathcal{C}_t}\ \bigcup_{s\in \mathcal{S}_c} D_t^{(c,s)}.
\]

Compared with conventional CIL, Stage-CIL requires more than inter-class discrimination. As learning proceeds, the model must retain recognition over the expanding label space $\mathcal{C}_{1:t}$ while preserving identity consistency across stage-specific forms of previously seen classes. This gives rise to \textbf{intra-class forgetting}: degraded recognition consistency across stages of the same semantic class.

\subsection{Benchmark Design and Evaluation}

We instantiate Stage-CIL through \textbf{Stage-Bench} under a controlled progressive protocol. In the current benchmark release, all classes share a common stage index set $\mathcal{S}=\{0,1\}$, where $s=0$ denotes an initial form and $s=1$ denotes a later evolved form. This two-stage design serves as the primary benchmark instantiation in this work. A configuration is denoted as $(\mathrm{B}\!-\!m,\,\mathrm{Inc}\!-\!n)\!\times\!S^2$, where an initial \emph{base} session introduces $m$ classes and each incremental session introduces $n$ new classes. For any class $c$, in the current Stage-Bench protocol, whenever both stages are available for a class, Stage-0 is exposed before Stage-1. The resulting stream consists of $B$ ordered learning steps. After each step $b\in\{1,\dots,B\}$, the model is evaluated on all classes and stages observed so far, yielding the Top-1 accuracy $\mathcal{A}_b$. The average incremental accuracy is then defined as $\bar{\mathcal{A}}=\frac{1}{B}\sum_{b=1}^{B}\mathcal{A}_b$. To quantify forgetting more finely, we report two complementary metrics. Let $N_{\mathrm{cls}}=|\mathcal{C}|$ denote the total number of classes in the benchmark, and let $A_i(b)$ be the accuracy of class $i$ at step $b$, aggregated over all stages of class $i$ that have been introduced by step $b$.

\noindent\textbf{Inter-class forgetting (Inter-F).}
Following \cite{ChaudhryRWalk18}, we define Inter-F as the average drop from the best-achieved accuracy of each class to its final accuracy:
\begin{equation}
\label{eq:interF}
F_{\mathrm{inter}}
=\frac{1}{N_{\mathrm{cls}}}\sum_{i=1}^{N_{\mathrm{cls}}}\Big(\max_{1\leq b\leq B}A_i(b)-A_i(B)\Big).
\end{equation}
A larger $F_{\mathrm{inter}}$ indicates more severe forgetting of previously learned classes in the conventional CIL sense.

\noindent\textbf{Intra-class forgetting (Intra-F).}
Standard CIL metrics do not reveal whether a model preserves recognition of the \emph{same class} after later-stage observations are introduced. To quantify this effect, we define Intra-F as a \emph{diagnostic metric} that measures the performance drop on the earliest morphological stage of each class, from its initial acquisition to the end of the full continual-learning stream. Let $b_i$ denote the evaluation step immediately after class $i$ is first learned at Stage-0, let $B$ denote the final evaluation step, and let $A_{i,0}^{(t)}$ denote the accuracy of class $i$ on its Stage-0 test subset evaluated at step $t$. Under the current two-stage protocol, we define
\begin{equation}
\label{eq:intraF}
F_{\mathrm{intra}}
=
\frac{1}{N_{\mathrm{cls}}}
\sum_{i=1}^{N_{\mathrm{cls}}}
\left[
\frac{A_{i,0}^{(b_i)}-A_{i,0}^{(B)}}
{\max(\epsilon, A_{i,0}^{(b_i)})}
\right]_+,
\qquad \epsilon = 10^{-6},
\end{equation}
where $[z]_+=\max(z,0)$. We use the positive-part operator to measure degradation only, without allowing improvements on some classes to offset forgetting on others. A larger $F_{\mathrm{intra}}$ indicates more severe forgetting of the earliest morphological stage of a class after subsequent continual updates. Unlike a terminal cross-stage accuracy gap, this metric compares the \emph{same} morphological stage before and after later-stage learning, and thus more directly reflects within-class forgetting under stage evolution. Although Stage-CIL naturally extends to more than two stages, the current benchmark primarily instantiates a two-stage regime, under which the above definition already captures the core challenge studied in this work. For more than two stages, the same before--after degradation is computed for every previously introduced non-terminal stage and then averaged across stages and classes.

\begin{figure}[t]
    \centering
    \begin{subfigure}{\columnwidth}
        \centering
        \includegraphics[width=\textwidth]{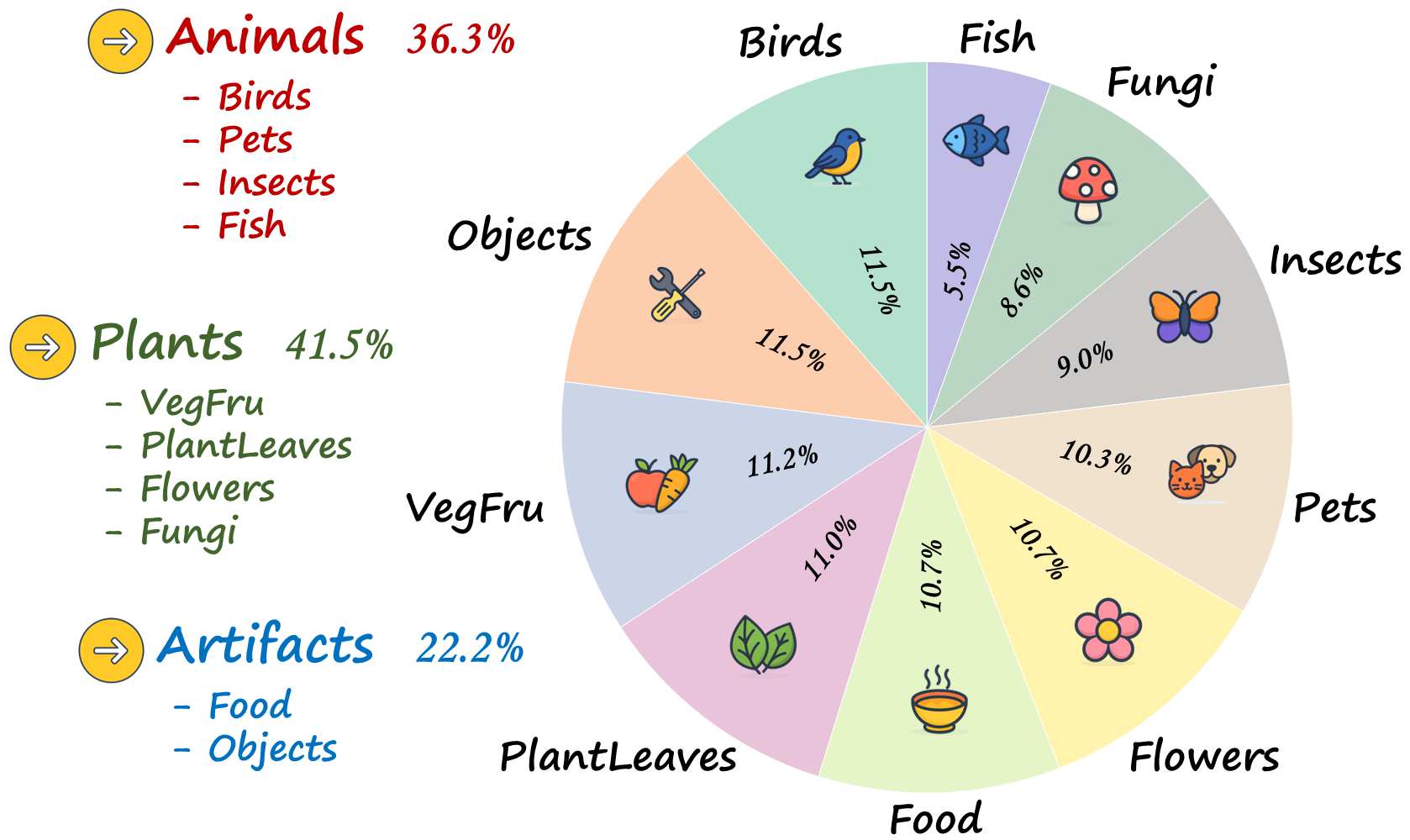}
        \caption{Semantic composition of Stage-Bench.}
    \end{subfigure}

    \vspace{0.5em}

    \begin{subfigure}{0.9\columnwidth}
        \centering
        \includegraphics[width=\textwidth]{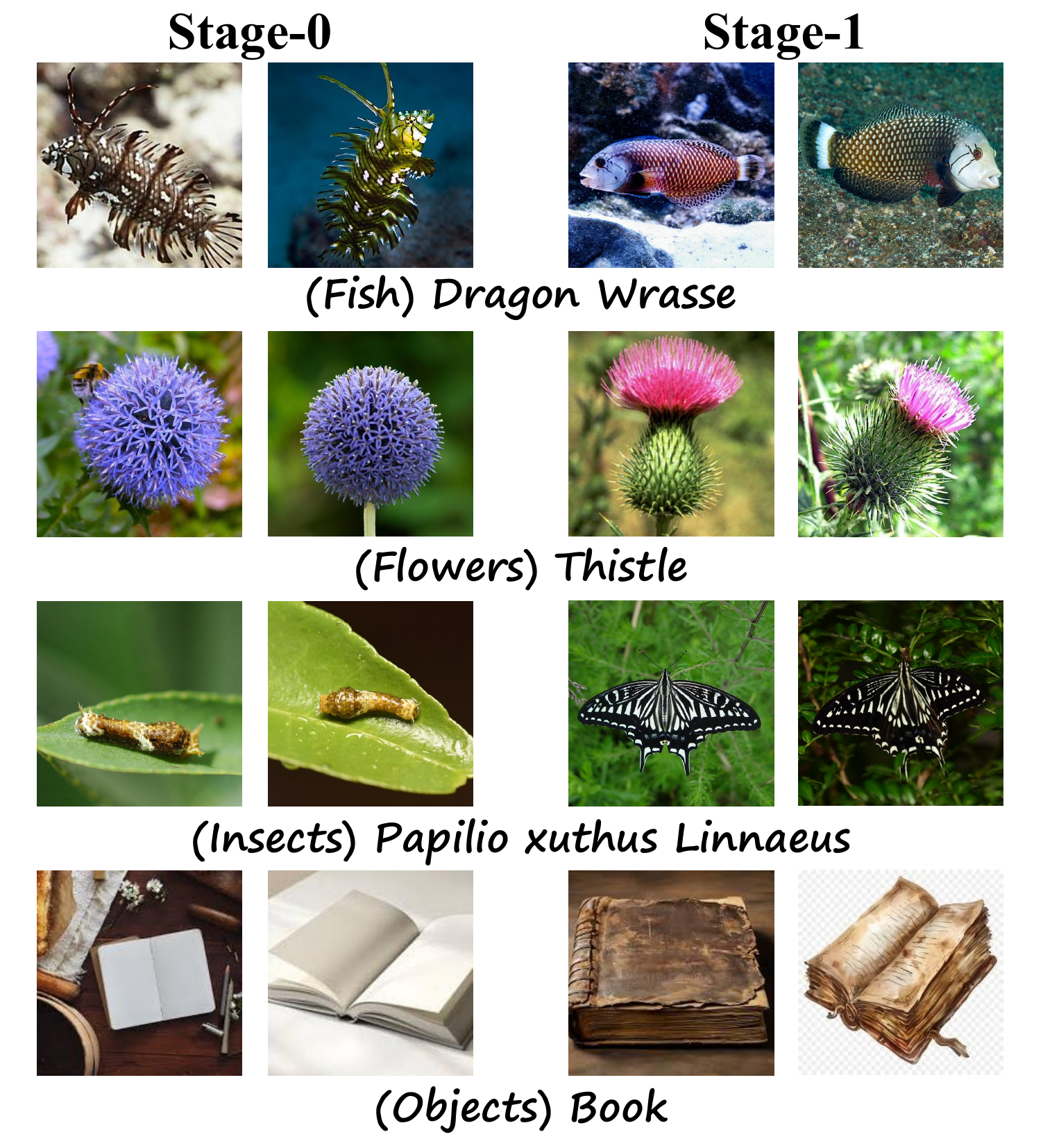}
        \caption{Representative Stage-0/1 samples from same classes.}
    \end{subfigure}

    \caption{Overview of \textbf{Stage-Bench}. (a) The benchmark spans 10 semantic domains, each with 20 classes and two ordered stages, yielding 400 morphological stages in total. (b) Representative examples show that stage variation is tied to the object itself.}
    \label{fig:stagebench_overview}
\end{figure}

To concretely instantiate Stage-CIL and its metrics, we construct and release \textbf{Stage-Bench}. The benchmark is designed around a simple principle: before moving to richer multi-stage settings, we first isolate the most pronounced form of intra-class evolution under a controlled protocol. In many real entities, the largest semantic and morphological shift occurs between an initial form and a clearly evolved form. Accordingly, the current release adopts a two-stage core regime (Stage-0 and Stage-1), which serves as a minimal non-trivial instantiation of Stage-CIL and provides a clean test bed for studying whether a learner can preserve semantic identity across pronounced stage-specific changes.

Stage-Bench spans $10$ diverse domains, with $20$ classes per domain and $18{,}895$ images in total. As summarized in Fig.~\ref{fig:stagebench_overview}, it covers a broad semantic spectrum across Animals, Plants, and Artifacts, while exhibiting semantically meaningful stage variation within classes. Each class is annotated with two ordered stages, and the benchmark provides standardized train/test splits together with a unified evaluation protocol. Importantly, the variation is driven by changes in the object itself, including growth, maturation, and morphological transformation, rather than merely by differences in background, style, or acquisition conditions. Further details on data collection, cleaning, and supplementary analyses are provided in the supplementary material.

\subsection{Stage-Oblivious Reduction}

A natural question is whether Stage-CIL can be simplified into a conventional class-incremental problem by discarding stage annotations and merging all samples of the same semantic class into a single label. To test this possibility, we construct a \emph{stage-oblivious reduction} of Stage-Bench: samples from different stages of each class are pooled together, yielding a standard $20$-step CIL stream with no explicit stage structure.

This reduction is easy to implement, but it changes the problem itself. Once stage information is removed, the learner is no longer asked to preserve identity coherence across evolving forms. Instead, it must compress substantially different appearances of the same class into a single static category. For evolving classes, such a reduction entangles semantic identity with large stage-specific variation, producing interference that conventional static-class CIL is not designed to resolve.

Table~\ref{tab:stage_oblivious_reduction} confirms that representative CIL baselines remain clearly limited under this stage-oblivious reduction. The purpose of this experiment is not to define a competing benchmark, but to test whether evolving classes can be faithfully collapsed into conventional CIL. The answer is negative: discarding stage structure weakens the formulation and fails to capture the core challenge of preserving semantic identity across evolving forms. This motivates evaluation under the proper Stage-CIL protocol.

\begin{table}[t]
\centering
\small
\caption{Average and final accuracy under the stage-oblivious reduction of Stage-Bench to conventional 20-step CIL. All methods are initialized with the same pre-trained CLIP for a fair comparison.}
\label{tab:stage_oblivious_reduction}
\setlength{\tabcolsep}{4pt}
\begin{tabular*}{0.8\columnwidth}{@{\extracolsep{\fill}}lcc@{}}
\toprule
Method & $\bar{\mathcal{A}}$ & $\mathcal{A}_B$ \\
\midrule

iCaRL ~(CVPR 2017) & ~~27.17~ & ~~15.53~ \\ DualPrompt ~(ECCV 2022) & ~~32.92~ & ~~19.16~ \\ CODA-Prompt ~(CVPR2023) & ~~33.84~ & ~~21.80~\\ EASE ~(CVPR2024) & ~~44.03~ & ~~31.22~\\ SimpleCIL ~(IJCV 2024) & ~~30.28~ & ~~18.77~\\ MOS ~(AAAI 2025) & ~~50.22~ & ~~36.34~ \\ PROOF ~(TPAMI 2025) & ~~45.65~ & ~~30.81~ \\ BOFA ~(AAAI 2026) & ~~47.58~ & ~~33.90~ \\
\bottomrule
\end{tabular*}
\end{table}

\begin{figure*}[ht]
  \centering
  \includegraphics[width=0.8\textwidth]{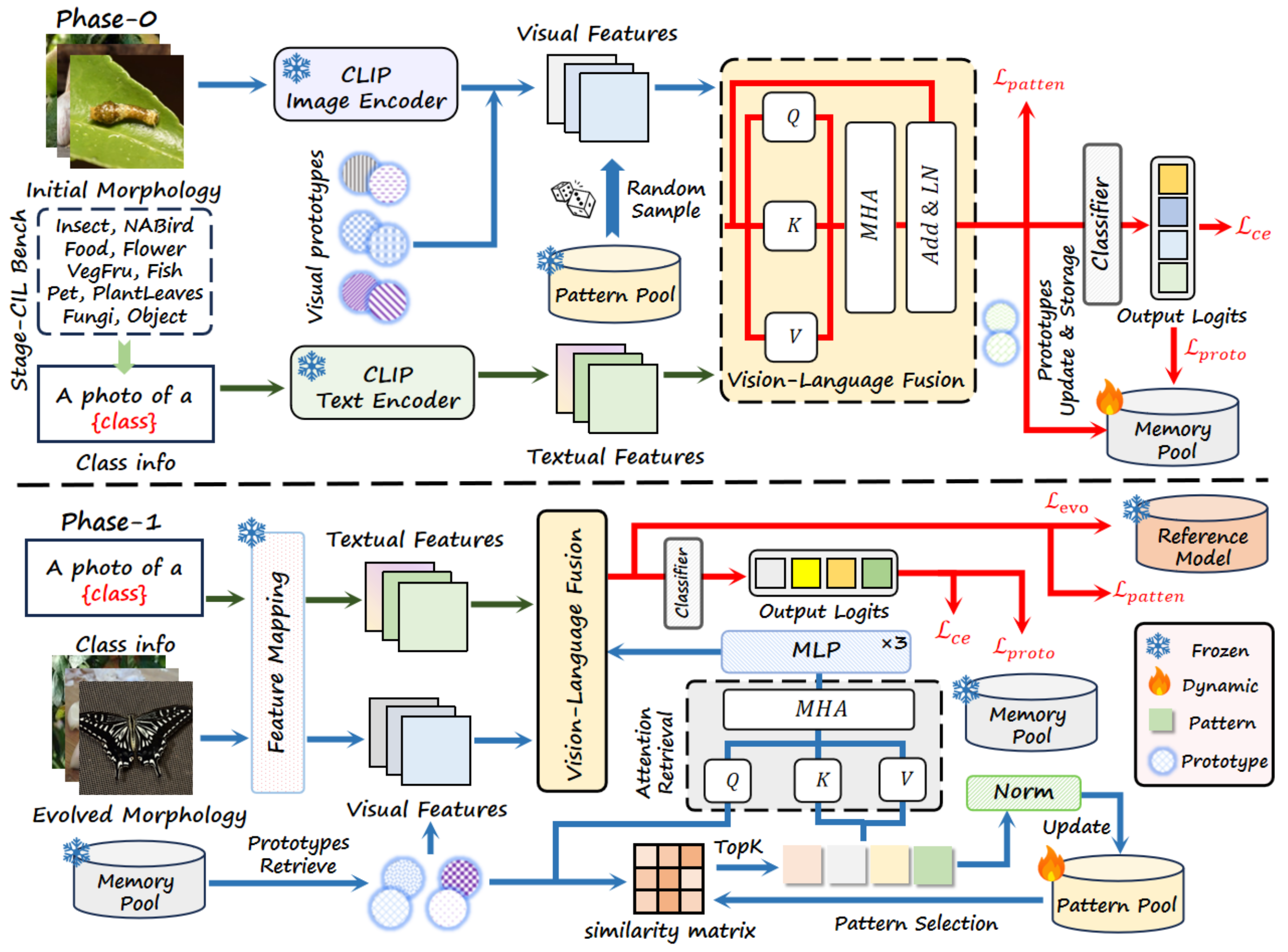}
  \caption{Illustration of STAGE. \textbf{Top: }the model learns a visual prototype for each new class from initial-morphology images, while text prompts provide class semantics. \textbf{Bottom: }the stored prototype queries the pattern pool, attends to the top-k patterns, and predicts the evolved representation, enabling classification of the Stage 1 images and online pattern updates.}
  \label{fig:framework}
\end{figure*}
\section{The STAGE Reference Baseline}

To demonstrate that Stage-CIL is learnable, we introduce \textbf{STAGE}, an evolution-aware reference baseline. STAGE follows a predict-then-classify procedure: it constructs a stable Stage 0 identity prototype and predicts its later-stage counterpart using reusable transformation patterns from an Evolution-aware Memory Pool. Fig.~\ref{fig:framework} and Alg.~\ref{alg:stage_training} summarize the pipeline.

\subsection{Predictive Evolution Framework}

Stage-Bench uses the two-stage index set $\mathcal S=\{0,1\}$. For task $T_t$, let $D_t^s=\{(x,y,s)\in D_t\}$ denote its Stage $s$ subset and $S_s^y=\{(x,y,s)\in D_t^s\}$ the sample set of class $y\in\mathcal C_t$ at that stage.

\noindent\textbf{Identity Anchor Construction (Phase 0).}
In Phase 0, STAGE constructs an identity prototype for the initial morphology of each class. Following PTM-based CIL~\cite{zhou2025learning}, we freeze the backbone $(g_{\mathrm{img}}, g_{\mathrm{text}})$ and introduce task-specific image and text projection heads, $\Pi_{\mathrm{img}}^t$ and $\Pi_{\mathrm{text}}^t$. Cross-modal fusion is learned with the first task and then fixed to establish a common prototype space. On a later task $T_t$, only its new heads are optimized on $D_t^0$, while previous heads remain frozen. Their outputs are accumulated without a task identifier:
\begin{align}
    \phi_{\mathrm{img}}^{(t)}(u) &= \sum_{m=1}^{t}\Pi_{\mathrm{img}}^m(u), &
    \phi_{\mathrm{text}}^{(t)}(u) &= \sum_{m=1}^{t}\Pi_{\mathrm{text}}^m(u). \label{eq:projection_path}
\end{align}
The accumulated image path encodes both training and test images, while the text path encodes the class prompt. For stage $s$, the class-mean visual representation is
\begin{align}
    v_s^y &= \frac{1}{|S_s^y|} \sum_{(x,y,s)\in S_s^y}
    \phi_{\mathrm{img}}^{(t)}\!\left(g_{\mathrm{img}}(x)\right). \label{eq:mean_proto}
\end{align}
Let $e_{\mathrm{text}}^y=\phi_{\mathrm{text}}^{(t)}(g_{\mathrm{text}}(\operatorname{prompt}(y)))$. Cross-modal attention fuses it with the visual mean in the common prediction space:
\begin{align}
    p_s^y &= W_q v_s^y + \mathrm{Attn}(W_q v_s^y,\, W_k e_{\mathrm{text}}^y,\, W_v e_{\mathrm{text}}^y). \label{eq:fused_proto}
\end{align}
In Phase 0, Eq.~\eqref{eq:fused_proto} produces the identity anchor $p_0^y$, stored as a fixed vector that subsequent projection updates neither re-encode nor modify. Fixed fusion and historical heads preserve the anchor space, while prototype rehearsal keeps the accumulated representation compatible with it. In Phase 1, all projection and fusion parameters are frozen, and the same equations construct the supervision target $p_1^y$ from $S_1^y$. Thus, $p_0^y$ and $p_1^y$ are class prototypes in the same space rather than paired instance features.

\begin{algorithm}[tb]
   \caption{STAGE Training Procedure}
   \label{alg:stage_training}
\begin{algorithmic}[1]
   \State {\bfseries Require:} Task sequence $T=\{T_1,\dots,T_N\}$; for each $T_t$: $D_t^0$, $D_t^1$; Pattern pool $\mathcal P$; Evolution net $E$
   \For{each task $T_t \in T$}
      \State {\bfseries Phase 0:}
      \State Train the current projection heads on $D_t^0$; train fusion only if $t=1$
      \State Compute and store $p_0^y$ via Eqs.~\eqref{eq:mean_proto}--\eqref{eq:fused_proto}

      \State {\bfseries Phase 1:}
      \State Freeze the projection and fusion parameters
      \State Compute target prototypes $p_1^y$ from $D_t^1$
      \For{each mini-batch $B \subset D_t^1$}
         \For{each class $y$ represented in $B$}
            \State Retrieve $p_0^y$ and select top-$k$ patterns from $\mathcal P$
            \State Predict $\hat{p}_1^y$ via Eq.~\eqref{eq:prediction}
         \EndFor
         \State Compute semantic-class logits via Eq.~\eqref{eq:classification}
         \State Compute evolution, memory-pool, and rehearsal losses
         \State Update the differentiable parameters via Eq.~\eqref{eq:loss_phase1}
         \State Apply Eq.~\eqref{eq:pattern_update} per represented class without gradients
      \EndFor
   \EndFor
\end{algorithmic}
\end{algorithm}

\noindent\textbf{Predictive Evolution via Memory Pool (Phase 1).}
In Phase 1, STAGE predicts the evolved class prototype from its Stage 0 anchor rather than replacing the anchor with mixed-stage features, separating transformation modeling from identity preservation. It uses an \textbf{Evolution-aware Memory Pool}, a fixed-size trainable memory $\mathcal P=\{r_1,\dots,r_K\}$ where each vector $r_i$ encodes a reusable transformation pattern. We reserve $p_s^y$ for the prototype of semantic class $y$ at stage $s$ and use $r_i$ exclusively for a pattern-pool vector.

Given $p_0^y$, the pool selects its cosine top-$k$ patterns as the keys and values of an attention module, with $p_0^y$ as the query. The module outputs a class-conditioned context $c^y$ and attention weights $w_i^y$:
\begin{align}
    c^y = \mathrm{Attn}\big(p_0^y,\{r_i\}_{i\in \mathrm{top}\text{-}k}\big). \label{eq:attention_context}
\end{align}
The evolution network then predicts the Stage 1 class prototype through a residual update:
\begin{align}
    \hat{p}_1^y = p_0^y + E(p_0^y, c^y). \label{eq:prediction}
\end{align}
The selected patterns are linearly composed within the active set, while attention and the nonlinear evolution network yield an anchor-conditioned predictor. This supports heterogeneous stage transitions with a shared, fixed-size pool.

\noindent\textbf{Stage-Agnostic Classification.}
Stage annotations organize training but are unavailable at inference. Let $z(x)=W_q\phi_{\mathrm{img}}^{(t)}(g_{\mathrm{img}}(x))$ denote a test image in the prototype space, and let $\mathcal A_y$ contain the available prototypes: $\{p_0^y\}$ after Phase 0 and $\{p_0^y,\hat{p}_1^y\}$ after Phase 1. Each semantic-class logit takes the best match across its stage prototypes,
\begin{align}
    \ell_y(x) &= \tau \max_{p\in\mathcal A_y}\cos\big(z(x),p\big),
    & \hat y &= \arg\max_{y\in\mathcal C_{1:t}}\ell_y(x), \label{eq:classification}
\end{align}
where $\tau$ is the logit scale. Neither prototype selection nor prediction uses a ground-truth stage label; cross-entropy uses these semantic-class logits.

\noindent\textbf{Online Pattern Adaptation.}
The memory pool is adapted through two explicit steps. First, the selected patterns receive gradients through the attention context and the training objective. Let $\widetilde r_i$ denote a selected pattern after this gradient step. We then apply a non-gradient competitive decay-and-injection update using the ground-truth class-level displacement $\Delta^y=p_1^y-p_0^y$:
\begin{equation}
    r_i \leftarrow (1-\eta)\,\widetilde r_i
    + \eta\, w_i^y\,\operatorname{stopgrad}\!\left(\Delta^y\right),
    \quad \forall i \in \mathrm{top}\text{-}k,
    \label{eq:pattern_update}
\end{equation}
where $w_i^y$ is the attention weight assigned to pattern $r_i$ for class $y$; patterns outside the selected set are unchanged by Eq.~\eqref{eq:pattern_update}. Unlike a standard weighted EMA, the retention coefficient here is shared across selected patterns, while $w_i^y$ controls only the injected displacement. This deliberate competition suppresses weakly matched selected patterns and reinforces patterns aligned with the observed class evolution. The evolution network $E(\cdot)$ is updated only by gradient descent, whereas the pool follows this hybrid gradient-plus-online update.

\subsection{Training Objectives}
The training objective is designed to jointly support accurate transformation modeling and robust knowledge retention in the Stage-CIL setting. During the initial stage (\textbf{Phase 0}), the model is optimized to learn a stable and discriminative representation for the initial morphology of each class through a composite objective:
\begin{equation}
    \mathcal L^{\mathrm{S0}} = \mathcal L_{\mathrm{ce}} + \gamma \mathcal L_{\mathrm{proto}} + \delta \mathcal L_{\mathrm{pattern}},
    \label{eq:loss_phase0}
\end{equation}
where $\mathcal L_{\mathrm{ce}}$ denotes the cross-entropy loss on the current batch, $\mathcal L_{\mathrm{proto}}$ rehearses stored class prototypes to keep the accumulated projection path aligned with the fixed anchor space, and $\mathcal L_{\mathrm{pattern}}$ preserves the evolution predictor on previously learned pattern states. Let
\begin{align}
    \mathcal F_{\mathcal P}(u)
    &= u + E\big(u,c(u;\mathcal P)\big), \nonumber\\
    c(u;\mathcal P)
    &= \mathrm{Attn}\big(u,\{r_j\}_{j\in\mathrm{top}\text{-}k(u)}\big)
    \label{eq:pattern_predictor}
\end{align}
denote the complete pattern-conditioned predictor. We randomly sample stored patterns as pseudo-input states and retain the predictor's identity response on them:
\begin{equation}
    \mathcal L_{\mathrm{pattern}}
    = \mathbb{E}_{r_i \sim \mathcal P}
    \left\|\mathcal F_{\mathcal P}\!\left(\operatorname{stopgrad}(r_i)\right)
    - \operatorname{stopgrad}(r_i)\right\|_2^2,
    \label{eq:loss_pattern}
\end{equation}
This objective is a predictor-consistency rehearsal term: it prevents later optimization from arbitrarily changing the evolution operator around representative states retained in the pool. It does not assume that a sampled pattern means ``no transformation,'' nor does it treat $r_i$ as a paired Stage 0/Stage 1 observation.

In the next stage (\textbf{Phase 1}), the model is required to predict the evolved morphology for the same class. To supervise this process, we introduce an evolution loss:
\begin{equation}
\begin{split}
    \mathcal L_{\mathrm{evo}} = \frac{1}{|\mathcal C_t|}\sum_{y\in\mathcal C_t}
    \left[\left\|\hat{p}_1^y-p_1^y\right\|_2^2
    + \lambda_{\mathrm{cos}}\Big(1-\cos(\hat{p}_1^y-p_0^y,\;p_1^y-p_0^y)\Big)\right],
\end{split}
\label{eq:loss_evo}
\end{equation}
where $\hat{p}_1^y$ is the predicted Stage 1 prototype and $p_1^y$ is the class-mean target defined by Eqs.~\eqref{eq:mean_proto}--\eqref{eq:fused_proto}. The first term reconstructs the evolved class prototype, while the cosine term aligns the predicted class-level displacement with its ground-truth direction. No paired Stage 0/Stage 1 instances are assumed.

The total objective in Phase 1 combines this evolution term with the same rehearsal components:
\begin{equation}
    \mathcal L^{\mathrm{S1}} = \mathcal L_{\mathrm{ce}} + \alpha \mathcal L_{\mathrm{evo}} + \gamma \mathcal L_{\mathrm{proto}} + \delta \mathcal L_{\mathrm{pattern}},
    \label{eq:loss_phase1}
\end{equation}
allowing the same predictive operator and shared memory pool to be reused throughout the incremental Stage-CIL process.

\section{Experiments}

\begin{table*}[t]
\caption{Average, last performance, and inter-/intra-forgetting comparison on Stage-Bench (\textbf{2}-stage) using CLIP with ViT-B/16 LAION-400M as the backbone. The best performance is shown in \textbf{bold}.}
\label{tab:com_stagebench}
\vskip 0.15in
\begin{center}
\setlength{\tabcolsep}{0pt}
\begin{tabular*}{0.85\textwidth}{@{\extracolsep{\fill}} lcccccc}
\toprule
& \multicolumn{6}{c}{\textbf{Stage-Bench (B-0 Inc-10)$\times$S$^2$}} \\
\cmidrule(lr){2-7}
\textbf{Method} & $\bar{\mathcal{A}}$ & $\mathcal{A}_B$ & $\mathcal{A}_{B,0}$ & $\mathcal{A}_{B,1}$ & Inter-F & Intra-F \\
\midrule
iCaRL ~(CVPR 2017)       & 47.26 & 33.12 & 25.41 & 32.89 & 24.31 & 36.42 \\
L2P ~(CVPR 2022)         & 41.05 & 27.16 & 21.96 & 32.43 & 26.77 & 51.08 \\
DualPrompt ~(ECCV 2022)  & 40.08 & 25.56 & 19.63 & 31.59 & 28.53 & 47.33 \\
CODA-Prompt ~(CVPR 2023) & 50.77 & 34.93 & 30.70 & 39.24 & 24.04 & 38.19 \\
EASE ~(CVPR 2024)        & 50.57 & 37.75 & 33.94 & 41.62 & 21.85 & 27.05 \\
SimpleCIL ~(IJCV 2025)   & 45.74 & 23.07 & 18.11 & 28.13 & 26.18 & 58.61 \\
MOS ~(AAAI 2025)         & 59.74 & 41.31 & 37.95 & 44.73 & 19.86 & 24.44 \\
PROOF ~(TPAMI 2025)      & 57.77 & 37.52 & 33.09 & 42.03 & 20.39 & 29.37 \\
BOFA ~(AAAI 2026)        & 58.60 & 37.64 & 35.63 & 44.11 & 20.27 & 26.68\\
S-Prompts ~(NeurIPS 2022)   & 44.32 & 29.18 & 22.47 & 35.89 & 25.64 & 42.76 \\
DCE ~(ICML 2025)         & 53.86 & 38.94 & 34.12 & 43.76 & 22.15 & 22.58 \\
\midrule
\textbf{STAGE} & \textbf{75.11} & \textbf{56.44} & \textbf{55.68} & \textbf{57.21} & \textbf{16.92} & \textbf{7.48} \\
\bottomrule
\end{tabular*}
\end{center}
\vskip -0.1in
\end{table*} 


\begin{table*}[t]
\centering
\caption{Average, last performance, and intra-forgetting comparison on the restructured Object domain using CLIP with ViT-B/16 LAION-400M as the backbone. The left part shows results under the 2-stage setting, and the right part shows results under the 3-stage setting.}
\label{tab:object_stage_compare}
\setlength{\tabcolsep}{0pt}
\begin{tabularx}{0.85\textwidth}{l *{6}{Y}}
\toprule
& \multicolumn{3}{c}{\textbf{Stage-Bench (Object) (B-0 Inc-10)$\times$S$^2$}}
& \multicolumn{3}{c}{\textbf{Stage-Bench (Object) (B-0 Inc-10)$\times$S$^3$}} \\
\cmidrule(lr){2-4} \cmidrule(lr){5-7}
\textbf{Method} & $\bar{\mathcal{A}}$ & $\mathcal{A}_B$ & Intra-F
& $\bar{\mathcal{A}}$ & $\mathcal{A}_B$ & Intra-F \\
\midrule
iCaRL ~(CVPR 2017)         & 96.84 & 96.20 & 1.44 & 95.83 & 94.67 & 1.52 \\
L2P ~(CVPR 2022)           & 94.84 & 93.18 & 3.14 & 93.42 & 92.84 & 3.67 \\
DualPrompt ~(ECCV 2022)    & 94.32 & 92.90 & 3.68 & 93.76 & 92.26 & 3.92 \\
CODA-Prompt ~(CVPR 2023)   & 96.76 & 95.33 & 2.82 & 96.38 & 95.42 & 3.35 \\
EASE ~(CVPR 2024)          & 98.73 & 98.04 & 1.37 & 96.52 & 95.18 & 1.38 \\
SimpleCIL ~(IJCV 2025)     & 97.59 & 95.94 & 3.46 & 94.25 & 93.91 & 3.28 \\
MOS ~(AAAI 2025)           & 98.92 & 98.33 & 1.43 & 97.42 & 97.25 & 1.68 \\
PROOF ~(TPAMI 2025)        & 97.84 & 96.93 & 1.81 & 97.15 & 96.21 & 2.07 \\
BOFA ~(AAAI 2026)           & 97.91 & 97.37 & 1.56 & 97.24 & 97.09 & 1.82 \\
S-Prompts ~(NeurIPS 2022)  & 95.06 & 94.25 & 2.08 & 94.02 & 93.15 & 2.35 \\
DCE ~(ICML 2025)           & 97.14 & 96.88 & 0.65 & 96.73 & 95.87 & 0.86 \\
\midrule
\textbf{STAGE}             & \textbf{99.21} & \textbf{99.12} & \textbf{0.22}
                          & \textbf{98.61} & \textbf{98.44} & \textbf{0.31} \\
\bottomrule
\end{tabularx}
\end{table*}

In this section, we validate the Stage-CIL paradigm and the Stage-Bench benchmark. We first establish the performance of various state-of-the-art (SOTA) CIL and DIL methods on this new benchmark, revealing the unique challenges posed by intra-class evolution. We then show STAGE effectively addresses these challenges and analyze its key components through ablation studies.

\subsection{Implementation Details}
 
\textbf{Dataset protocol. } For all experiments, we follow the protocol of the Stage-Bench described in metrics, using the (B-0, Inc-10) $\times$ S$^2$ protocol of Stage-Bench (400 morphological stages, 200 classes $\times$ 2 stages), which unfolds over 20 incremental steps and a total of 40 tasks. All experiments use a fixed random seed of 1993 for class order shuffling, and all methods are evaluated on the identical sequence of tasks and data splits for fair comparison, following \cite{zhou2025revisiting}. Additionally, to test the multi-stage generalization of STAGE, we restructured the Object domain into a three-stage evolution task, evaluated under the same protocol.

\noindent\textbf{Training Details. } We use Pytorch \cite{paszke2019pytorch} and PILOT \cite{SunPilot25} to implement all models on two NVIDIA TITAN X GPUs. We use the \emph{same} network backbone, \emph{i.e.}, CLIP with ViT-B/16 (OpenCLIP LAION-400M) \cite{IlharcoOpenCLIP21} for all compared methods for \emph{fair comparison}. We set the batch size to 16 and train for 5 epochs using SGD with momentum for optimization. The learning rate starts 0.001 and decays with cosine annealing. In practice, we interleave classification and evolution updates in Phase-1, rehearse a small, uniformly random subset of the memory pool per step, and use a small EMA rate for stable adaptation.

\noindent \textbf{Comparison methods. } We evaluate STAGE against a comprehensive suite of state-of-the-art incremental learning methods. We select prominent Pre-Trained Model-based approaches, including prompt-tuning methods like L2P \cite{wang2022learning}, DualPrompt \cite{wang2022dualprompt} and CODA-Prompt \cite{smith2023coda}, as well as other superior methods such as EASE \cite{zhou2024expandable}, SimpleCIL \cite{zhou2025revisiting}, MOS \cite{sun2025mos}, PROOF \cite{zhou2025learning} and BOFA \cite{li2026bofa}. Furthermore, to represent the perspective of DIL, which views stage evolution as a sequence of domain shifts, we include leading methods such as S-Prompts \cite{Wang2022SsPrompts} and DCE \cite{Li2025DCE}. In addition, we compare STAGE with a classic rehearsal-based method, iCaRL \cite{rebuffi2017icarl}, which we adapt for a PTM-based framework. To ensure a fair and rigorous comparison, all methods are implemented using the same PTM backbone and experimental protocol.

\begin{figure}[t]
  \centering
  \begin{subfigure}[b]{1\columnwidth}
    \centering
    \includegraphics[width=\linewidth]{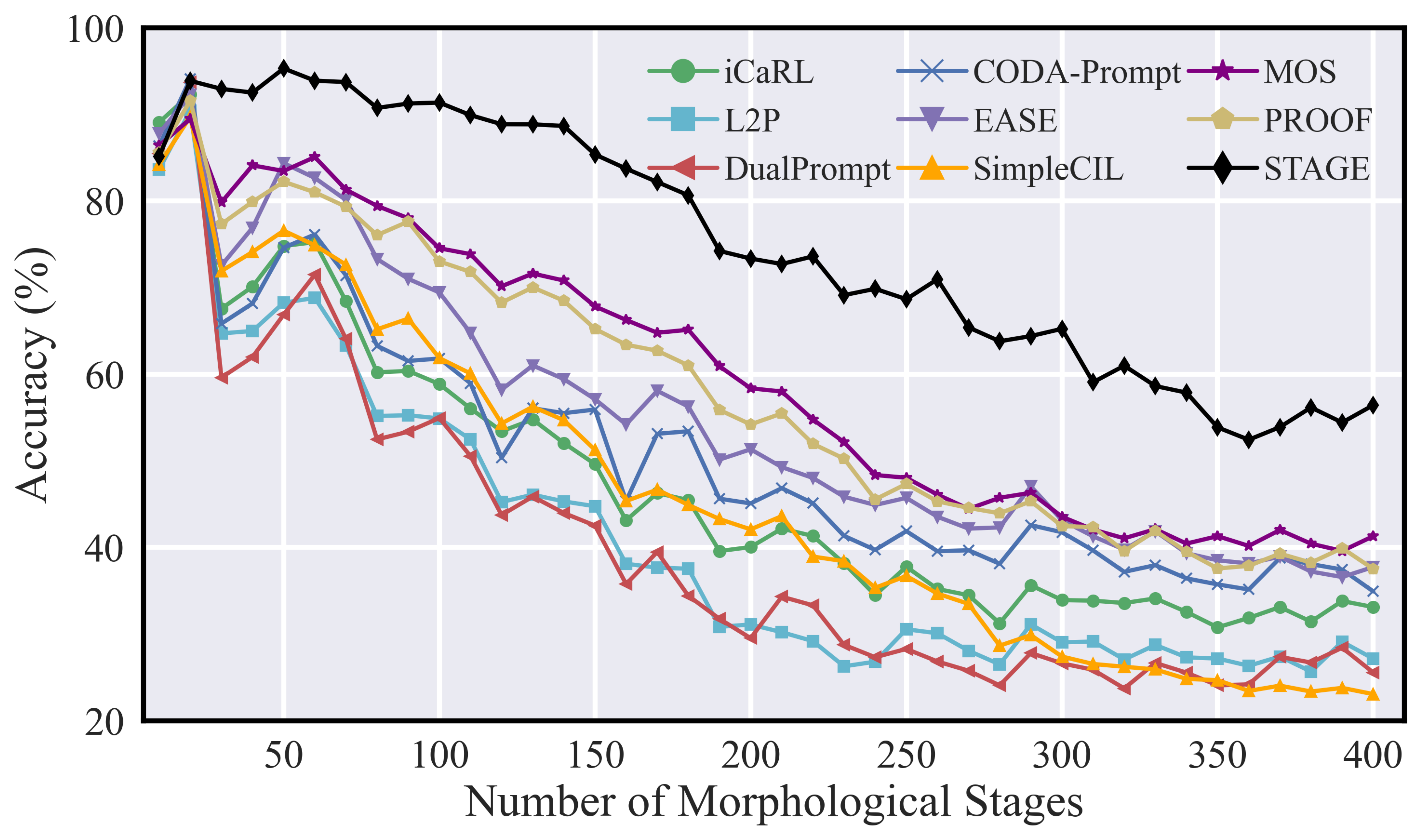}
    \label{fig:stage10-s2}
  \end{subfigure}
  \caption{Incremental performance of different methods. We report the performance gap after the last incremental stage of STAGE and the runner‑up method at the end of the line. All methods are based on the same backbone/weight.}
  \label{fig:result40}
\end{figure}

\subsection{Benchmark Comparison}

In this section, we evaluate STAGE against representative SOTA methods on Stage-Bench. The complete results, including final and average accuracy, are reported in Table~\ref{tab:com_stagebench}. On the primary two-stage benchmark, a clear performance hierarchy emerges: existing methods designed mainly for inter-class knowledge isolation remain clearly limited under the pronounced morphological shifts of Stage-CIL, as reflected by their high Inter-F and often even larger Intra-F values. In contrast, STAGE consistently delivers the strongest performance across the primary continual-learning metrics.
As shown in Figure~\ref{fig:result40}, STAGE establishes an advantage from the early tasks, and the gap further widens as more classes and stages are introduced, eventually exceeding 15\% in final accuracy near the end of the stream. This indicates stronger robustness under long-sequence class evolution.
Furthermore, on the restructured three-stage Object domain in Table~\ref{tab:object_stage_compare}, STAGE still maintains a clear lead and achieves a remarkably low Intra-F of 0.31\%, suggesting that explicit evolution-aware modeling remains effective beyond the two-stage core regime.

\subsection{Ablation Study}

\begin{figure}[t]
  \centering
  \begin{subfigure}[b]{1\columnwidth}
    \centering
    \includegraphics[width=\linewidth]{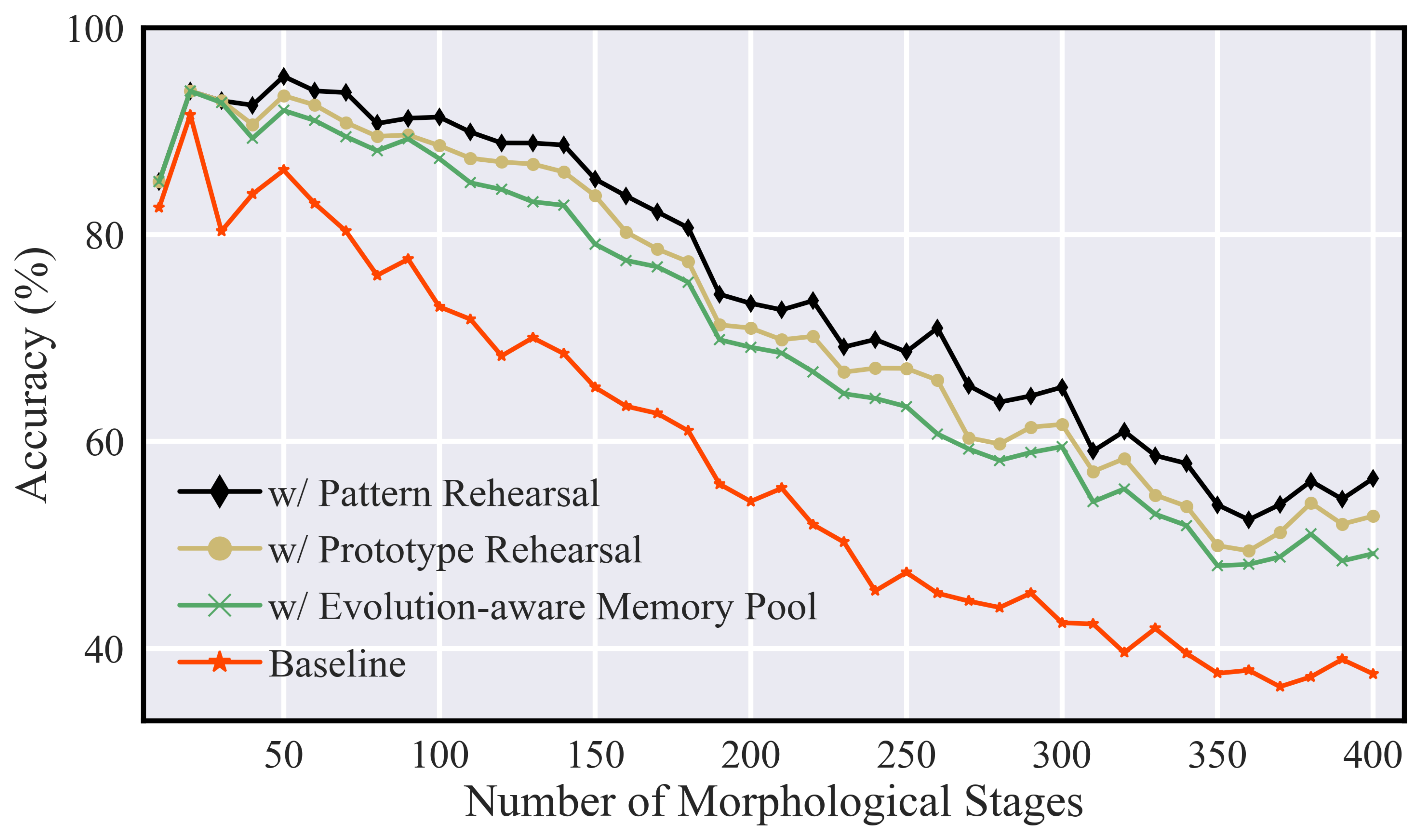}
    \label{fig:Ablation study}
  \end{subfigure}
  \caption{ablation of the STAGE on Stage-Bench. Each component contributes to either stage-conditioned prediction or continual retention under Stage-CIL.}
  \label{fig:result_comparison}
\end{figure}

To understand why the STAGE reference baseline remains effective under Stage-CIL, we conduct an ablation study, with results shown in Figure~\ref{fig:result_comparison}. The Baseline model, which lacks explicit evolution-aware modeling, performs poorly, indicating that conventional classification alone is insufficient when the same class appears in substantially different stages. The largest gain comes from introducing the ``w/ Evolution-aware Memory Pool'', suggesting that explicit stage-conditioned prediction is important for handling the feature drift caused by intra-class evolution. Adding ``w/ Prototype Rehearsal'' provides a further stable improvement, showing that conventional inter-class retention remains necessary under Stage-CIL. Finally, ``w/ Pattern Rehearsal'' offers an additional boost by regularizing the shared transformation memory. Overall, the results suggest that the reference baseline works by combining transition modeling with continual retention.

\begin{table}[t]
  \centering
  \caption{Sensitivity analysis of the STAGE on Stage-Bench under different pattern pool sizes and top-$k$ selections. We report average accuracy, final accuracy, inter-class forgetting, and intra-class forgetting.}

  \setlength{\tabcolsep}{5pt}
  \begin{tabular}{lcccc}
    \toprule
    \multicolumn{5}{c}{\textbf{Stage-Bench\quad (B-0\,Inc-10)$\times$S$^2$}} \\
    \midrule
    \textbf{Configuration} & $\bar{\mathcal{A}}$ & $\mathcal{A}_{B}$ & Inter-F & Intra-F \\
    \midrule
    \multicolumn{5}{c}{\textbf{Pattern Pool Size (Top-5)}} \\
    Pattern 10  & 71.15 & 47.60 & 19.48 & 15.73 \\
    Pattern 50  & 75.11 & 56.44 & 16.92 & 7.48 \\
    Pattern 100 & 76.12 & 57.38 & 16.80 & 6.83 \\
    \midrule
    \multicolumn{5}{c}{\textbf{Top-$k$ Selection (Pool Size 50)}} \\
    Top-1  & 71.13 & 47.29 & 19.84 & 18.05 \\
    Top-5  & 75.11 & 56.44 & 16.92 & 7.48 \\
    Top-10 & 73.20 & 52.31 & 17.83 & 9.16 \\
    \bottomrule
  \end{tabular}
  \label{tab:param_analysis}
\end{table}

Table~\ref{tab:param_analysis} examines pattern-pool size and top-$k$ selection. A pool of 50 with $k=5$ gives the best trade-off: 10 patterns provide insufficient capacity, whereas 100 bring little benefit. For retrieval, $k=1$ sharply increases Intra-F, while $k=10$ slightly reduces performance, indicating that $k=5$ balances transformation coverage against interference from less relevant patterns.

\section{Conclusion}

We introduced Stage-Aware Class-Incremental Learning (Stage-CIL), a setting for learning classes undergoing morphological evolution, and Stage-Bench, a 10-domain benchmark evaluating both inter- and intra-class forgetting. Experiments reveal the limitations of existing CIL and DIL methods under stage evolution. We further presented STAGE, an evolution-aware reference baseline that separates stable class identity from stage-dependent transformations and consistently improves performance. 
\begin{acks}
This work was supported by the Central-Oriented Foundation for Local Science and Technology Development, China.\\
\mbox{(No.~202407a12020010).}
\end{acks}

\bibliographystyle{ACM-Reference-Format}
\bibliography{icml}
\ifdefined\STAGEStandaloneSupplement
\else
  \clearpage
  \makeatletter
  \global\@ACM@balancefalse
  \makeatother
\fi

\appendix
\ifdefined\STAGEStandaloneSupplement
\else
  \section*{Supplementary Material}
\fi

This supplementary material provides additional details and analyses for the Stage-CIL benchmark and the STAGE reference baseline. It is organized as follows:
\begin{itemize}
    \item \textbf{Details of Stage-Bench (Appendix A)} describes the benchmark composition, data sources, stage annotation procedure, licensing and ethical considerations, and representative Stage-0/Stage-1 pairs.

    \item \textbf{Compared Methods (Appendix B)} summarizes the CIL and DIL baselines and discusses how their design assumptions differ from the Stage-CIL setting.

    \item \textbf{Additional Analysis of STAGE (Appendix C)} examines prototype stability, the dynamics of the shared pattern pool, and the computational and memory costs beyond the frozen CLIP backbone.

    \item \textbf{Additional Experimental Results (Appendix D)} reports diagnostic stage-aware controls, class-order robustness, per-domain performance, and qualitative visualizations.

    \item \textbf{Generalized Stage-CIL Protocol (Appendix E)} outlines a formulation beyond the two-stage case and defines an Intra-F diagnostic for intermediate stages.
\end{itemize}

The accompanying artifact release will include the source code, experiment scripts, and Stage-Bench construction tools.

\section{Details of Stage-Bench}
\begin{figure*}[!tbp]
  \centering
  \includegraphics[width=0.83\textwidth]{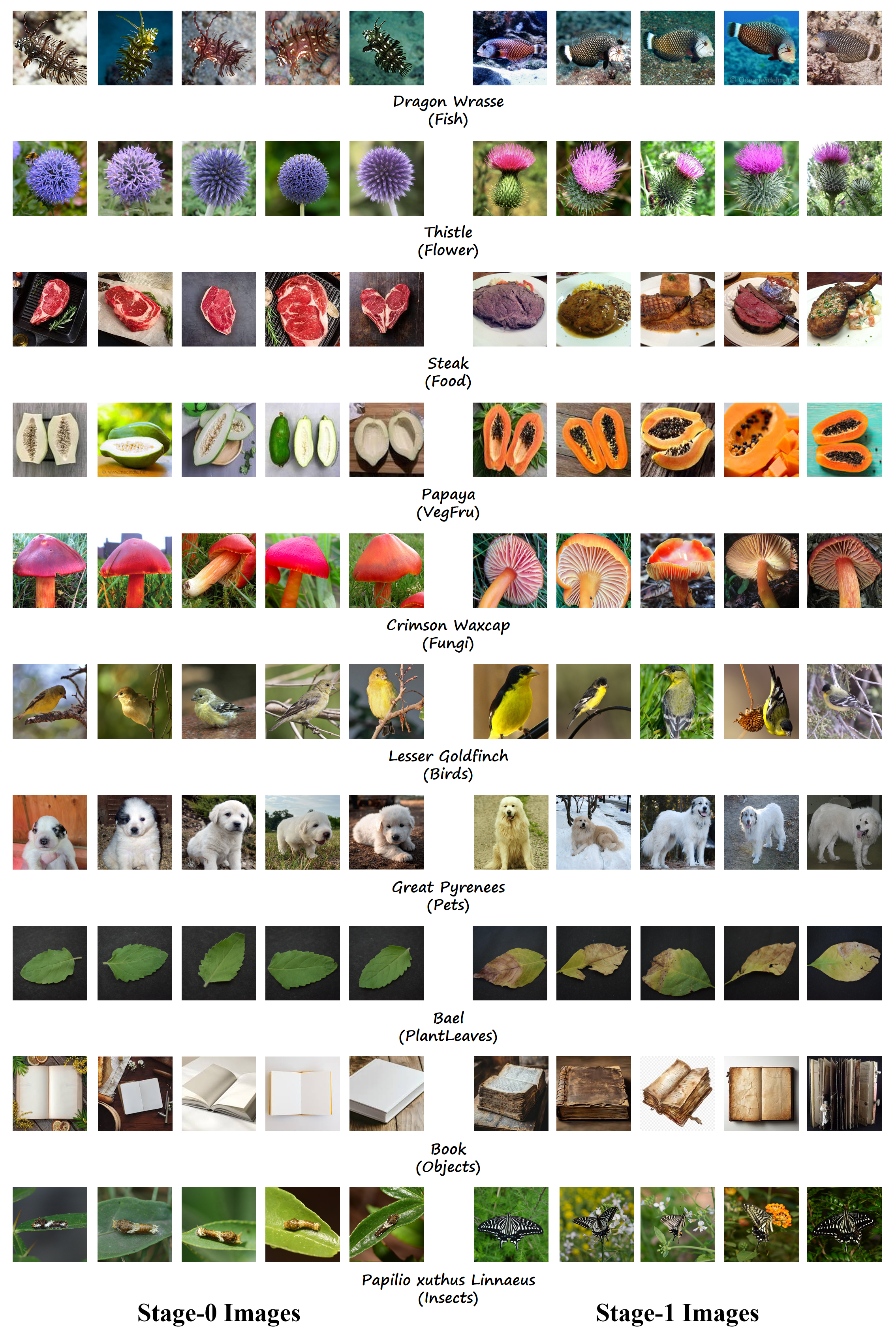}
  \caption{Illustration of Stage-Bench. Each row shows one class across two morphological stages: Stage-0 (left block, initial form) and Stage-1 (right block, evolved form).}
  \label{fig:Bench}
\end{figure*}

As summarized in the main paper, \textbf{Stage-Bench} spans ten domains, each containing 20 classes, with every class annotated with two ordered morphological stages: Stage-0 (initial) and Stage-1 (evolved). The benchmark comprises 18,895 images with a roughly $1\!:\!1$ balance between stages, yielding 400 morphological stages. Fig.~\ref{fig:Bench} provides concrete examples of stage definitions across the ten domains.

\paragraph{Why Preserve Cross-Stage Identity?}
Stage-CIL keeps the prediction target at the semantic class level: the stage annotation describes the observed morphology but does not turn each class--stage pair into an independent category. Merging all stages into one static training distribution removes the ordered transition structure, whereas treating $(y,s)$ as a pseudo-class explicitly pushes different stages of the same class apart. The latter is unsuitable when downstream decisions operate on persistent semantic identity, such as species monitoring across life stages, asset recognition across aging or damage states, and long-term retrieval of evolving entities.

\paragraph{Controlled Two-Stage Scope.}
The current release deliberately adopts a clean Stage-0$\to$Stage-1 protocol as the minimal non-trivial setting that isolates the conflict between stable semantic identity and pronounced morphological change. Stage annotations and temporal order are therefore controlled benchmark variables rather than quantities inferred by the learner. Noisy, missing, or partially ordered stage information constitutes a broader robustness setting and is not assumed by the results reported here.

\paragraph{Data Sources and Web Collection.}
A majority of images in Stage-Bench are drawn from established public benchmarks (e.g., Food-101~\cite{BossardFood14}, VegFru~\cite{Hou2017ICCV}, Flowers~\cite{nilsback2008automated}, Birdsnap~\cite{van2015building}, Pets~\cite{ParkhiPets12}, MVTec-AD~\cite{Bergmann2019MvtecAD}). For each domain, we first identified classes whose semantics naturally support a two-stage interpretation. When the original datasets lacked explicit stage labels, we constructed Stage-0 and Stage-1 using a combination of programmatic filtering and manual grouping of visually compatible subclasses. To increase diversity and balance classes, approximately 29.6\% of the images were additionally collected from publicly accessible web pages using class- and stage-specific keyword queries. Images and annotations for the Insects domain were sourced from our private internal database. This subset will be made available to qualified researchers for non-commercial research use. Access will be provided upon formal request and approval, after which a private download link will be issued.

\paragraph{Stage Annotation and Quality Control.}
Stage labels were assigned in a rigorous two-step process. First, an initial annotator proposed a coarse split into Stage-0 and Stage-1 based on salient morphological changes. Second, a different annotator independently reviewed all images for each class–stage pair to confirm, correct, or remove samples. Disagreements were resolved by a senior annotator. This quality-control loop was guided by a detailed annotation guideline containing textual descriptions and visual examples for each stage to promote consistent labeling decisions. Any image that remained ambiguous was discarded. This procedure ensures that the Stage-0/Stage-1 split reflects clear, semantically meaningful transitions. We will release mapping files linking Stage-Bench images to their original dataset identifiers or source URLs, enabling full traceability.

\paragraph{Licensing and Ethical Considerations.}
Stage-Bench is intended solely for non-commercial research and educational use. Images from existing benchmarks inherit their original licenses; we provide download scripts rather than redistributing the images themselves. For web-collected images, we only used content that was publicly accessible and manually filtered all images to remove personally identifiable information (PII). No personal, medical, or otherwise sensitive attributes are annotated. We will fully document data sources and usage constraints, and we ask downstream users to respect the original terms of use of all underlying datasets.

\section{Compared Methods}

We instantiate every comparison method with the same pre-trained backbone and evaluation protocol. The descriptions below summarize each method and identify the architectural mismatch evaluated by Stage-CIL; they do not imply that the original method cannot be redesigned for evolving classes.
\begin{itemize}
  \item \textbf{iCaRL}\cite{rebuffi2017icarl}: employs knowledge distillation, exemplar replay, and a nearest-mean-of-exemplars classifier. Its exemplars preserve selected observations, but its objective does not explicitly connect the ordered stages of one semantic class.

  \item \textbf{L2P}\cite{wang2022learning}: maintains a prompt pool for a pre-trained Vision Transformer and retrieves prompts for each instance by key--query similarity. Its prompt-selection objective does not explicitly preserve identity across ordered morphological stages.

  \item \textbf{DualPrompt}\cite{wang2022dualprompt}: augments L2P with general and expert prompts to represent task-invariant and task-specific information. The separation between prompt types mitigates task interference but does not impose continuity between stages of the same class.

  \item \textbf{CODA-Prompt}\cite{smith2023coda}: decomposes prompts into a bank of basis components and uses attention-based reweighting to compose instance-specific prompts. This query-dependent composition is not explicitly conditioned on stage order or trained to represent cross-stage transformations.

  \item \textbf{EASE}\cite{zhou2024expandable}: trains a lightweight adapter for each task and uses semantic-guided prototype completion to represent old classes in new subspaces. Its similarity-based completion does not explicitly encode the direction of morphological evolution.

  \item \textbf{SimpleCIL} \cite{zhou2025revisiting}: constructs a cosine classifier from class-wise prototypes extracted by a frozen PTM. In our Stage-CIL instantiation, updating a semantic class from its currently observed stage provides no separate mechanism for retaining and linking its earlier morphology.

  \item \textbf{MOS} \cite{sun2025mos}: introduces task-specific adapters, adapter merging, and self-refined adapter retrieval for PTM-based CIL. Its task retrieval and merging mechanisms do not explicitly model the ordered relation between class stages.

  \item \textbf{PROOF} \cite{zhou2025learning}: freezes the vision--language encoders and incrementally adds task-specific projection heads and a projection-fusion module. These components separate tasks effectively, but they do not explicitly predict how a semantic class changes between stages.

  \item \textbf{S-Prompts} \cite{Wang2022SsPrompts}: learns domain-specific prompts and retrieves one for each instance by feature similarity. Treating stages as domains captures domain selection, but does not explicitly exploit their ordered relation within a semantic class.

  \item \textbf{DCE} \cite{Li2025DCE}: uses frequency-aware expert networks and a dynamic selector for imbalanced domain-incremental learning. When stages are treated as domains, expert routing does not by itself enforce a shared identity or an explicit transition between stages.
\end{itemize}

\paragraph{Fairness and Difference from PROOF.}
All compared methods use the same frozen OpenCLIP ViT-B/16 backbone and pre-trained weights, so they begin with the same pre-trained knowledge. PROOF incrementally introduces task-specific projection heads and fuses their outputs, whereas STAGE stores a fixed identity anchor and predicts cross-stage residual transformations through one shared evolution-pattern pool. The evolution predictor operates on encoded feature vectors. Our implementation uses CLIP-based visual--textual fusion to construct semantic anchors, but the pattern-pool transition neither retrieves nor updates text tokens.

\section{Additional Analysis of STAGE}

\subsection{Stored-Prototype Stability and Retrieval Semantics}
The Stage-0 prototype of each class is stored after identity anchor construction and is not re-encoded by projection modules learned for later tasks. The cross-modal fusion module and historical projection heads remain frozen, while prototype rehearsal keeps newly accumulated projection outputs compatible with the fixed anchor space. This separates representational drift of a stored anchor from the morphological change that STAGE is designed to model. As a diagnostic, we recomputed the corresponding Stage-0 anchors after the continual stream and measured an average cosine similarity of $0.9982$ to the stored anchors, indicating that the identity reference remains stable.

The pattern pool retrieves residual transformation directions rather than absolute Stage-1 appearances. Specifically, each selected pattern contributes to predicting the change from the stored Stage-0 anchor to its later-stage representation. Consequently, retrieval is not a search for a later-stage image that visually resembles the initial morphology; the shared pool instead composes transformations learned across classes and domains.

\begin{figure}[!tbp]
  \centering
  \includegraphics[width=\columnwidth]{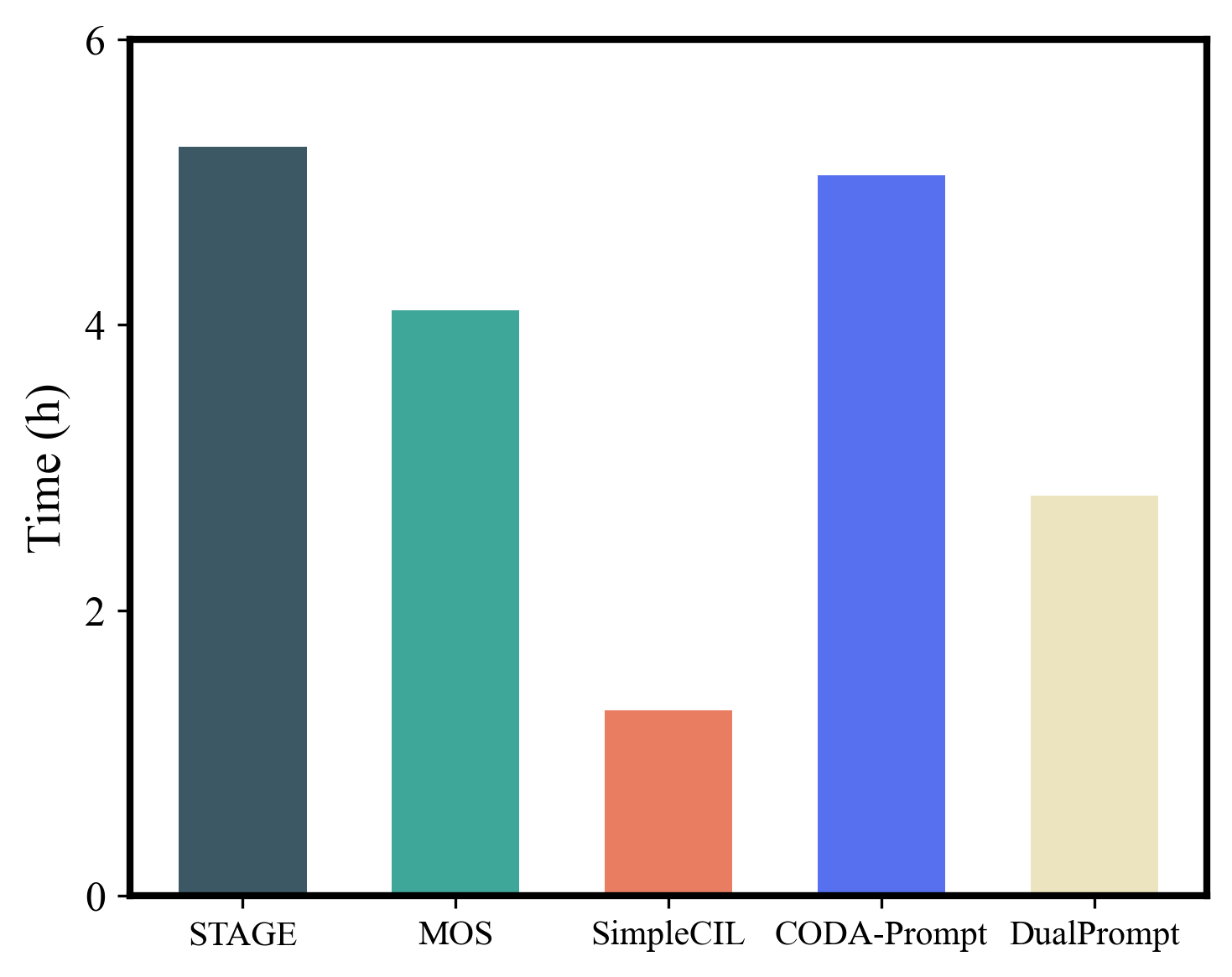}
  \caption{Running time comparison of different methods over Stage-Bench.}
  \label{fig:time}
\end{figure}

\begin{figure*}[!tbp]
  \centering
  \begin{subfigure}[b]{0.19\textwidth}
    \centering
    \includegraphics[width=\linewidth]{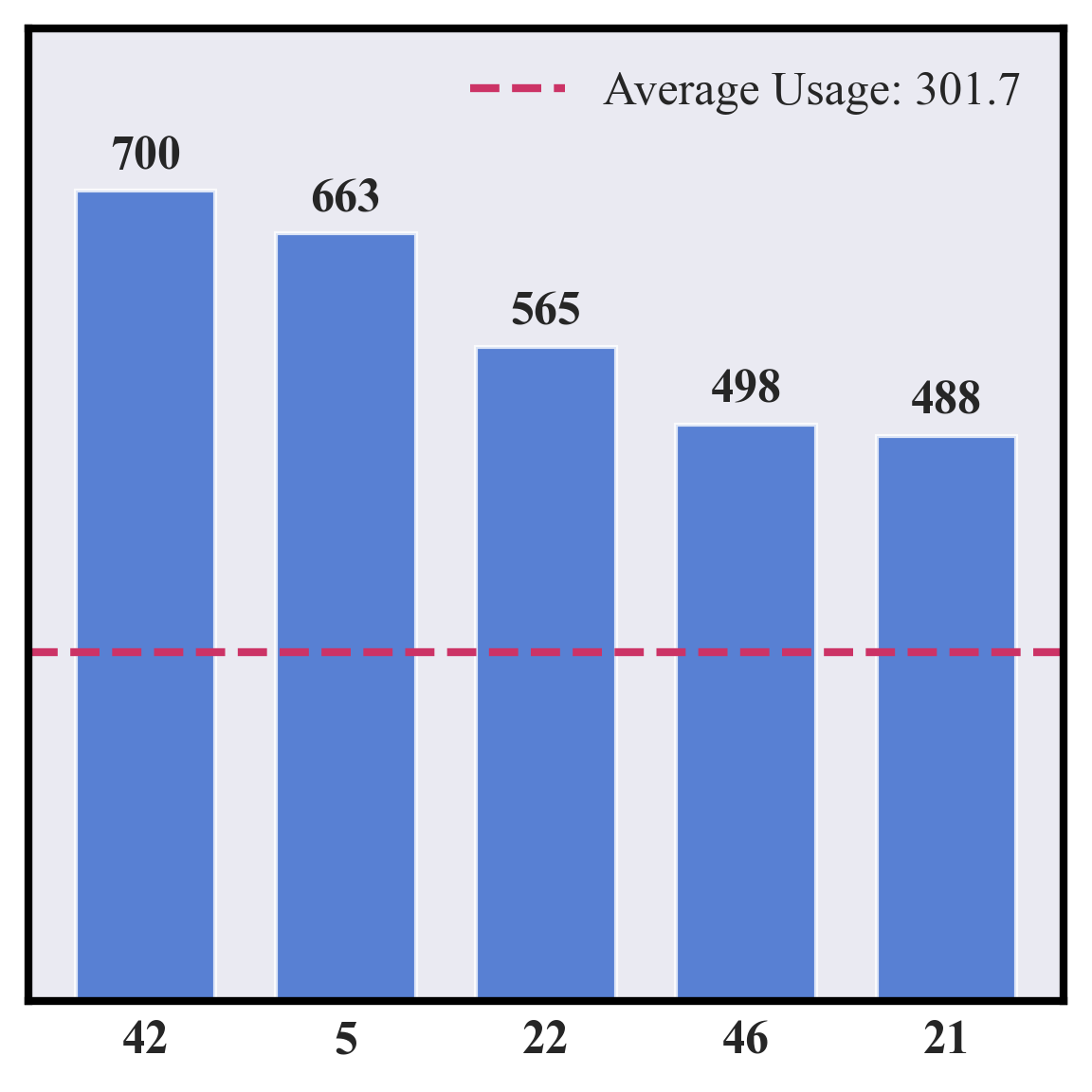}
    \caption{Domain 1}
  \end{subfigure}
  \begin{subfigure}[b]{0.19\textwidth}
    \centering
    \includegraphics[width=\linewidth]{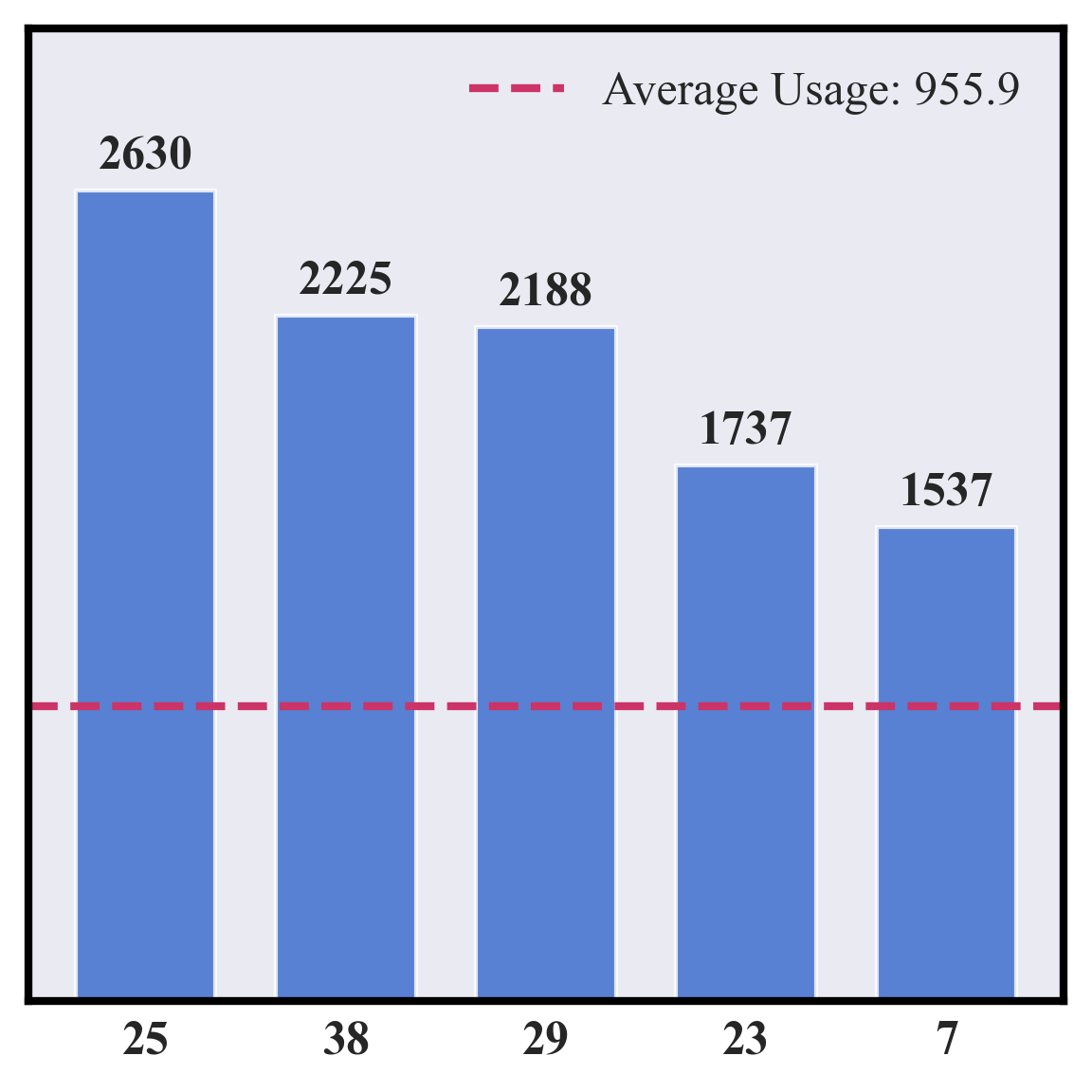}
    \caption{Domain 2}
  \end{subfigure}
  \begin{subfigure}[b]{0.19\textwidth}
    \centering
    \includegraphics[width=\linewidth]{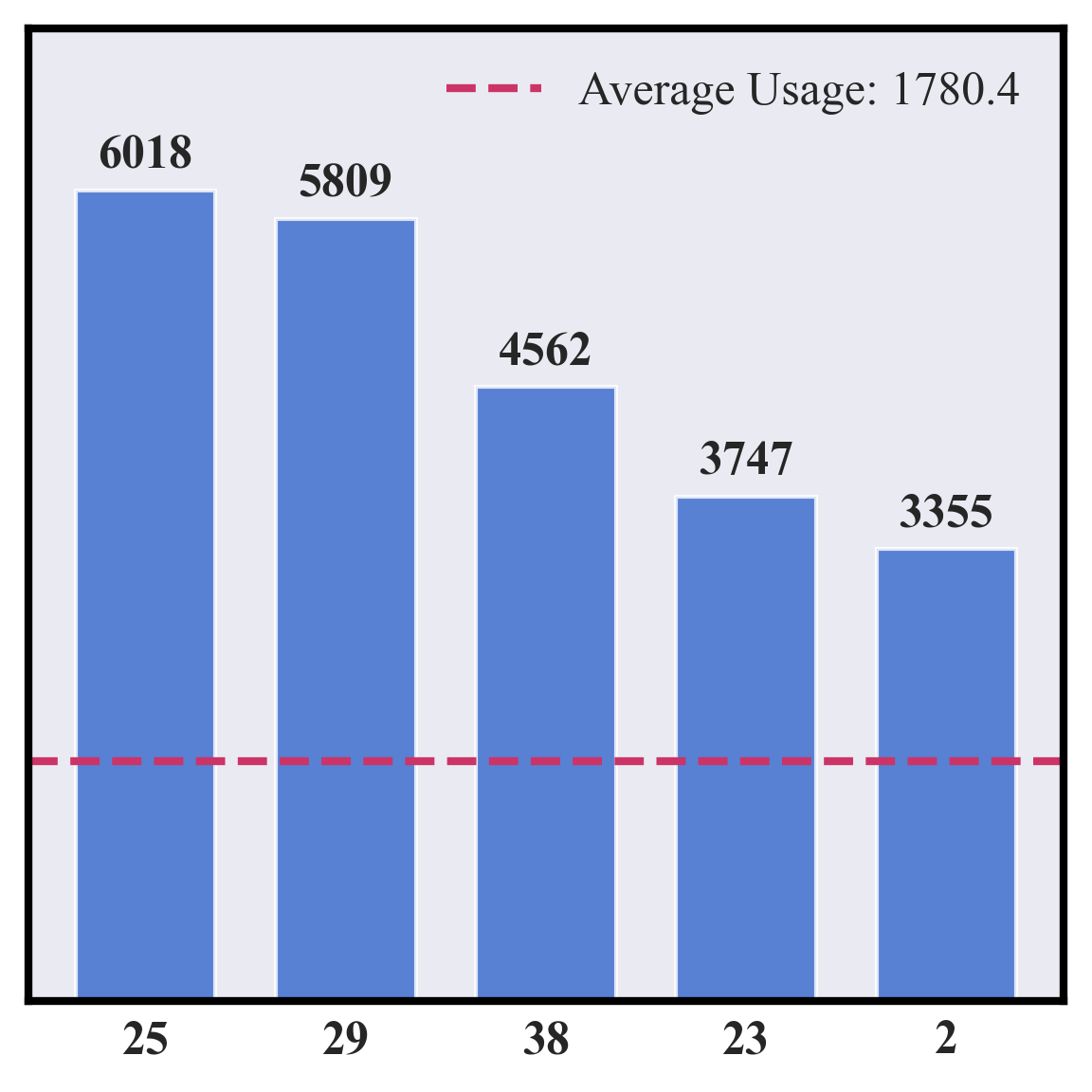}
    \caption{Domain 3}
  \end{subfigure}
  \begin{subfigure}[b]{0.19\textwidth}
    \centering
    \includegraphics[width=\linewidth]{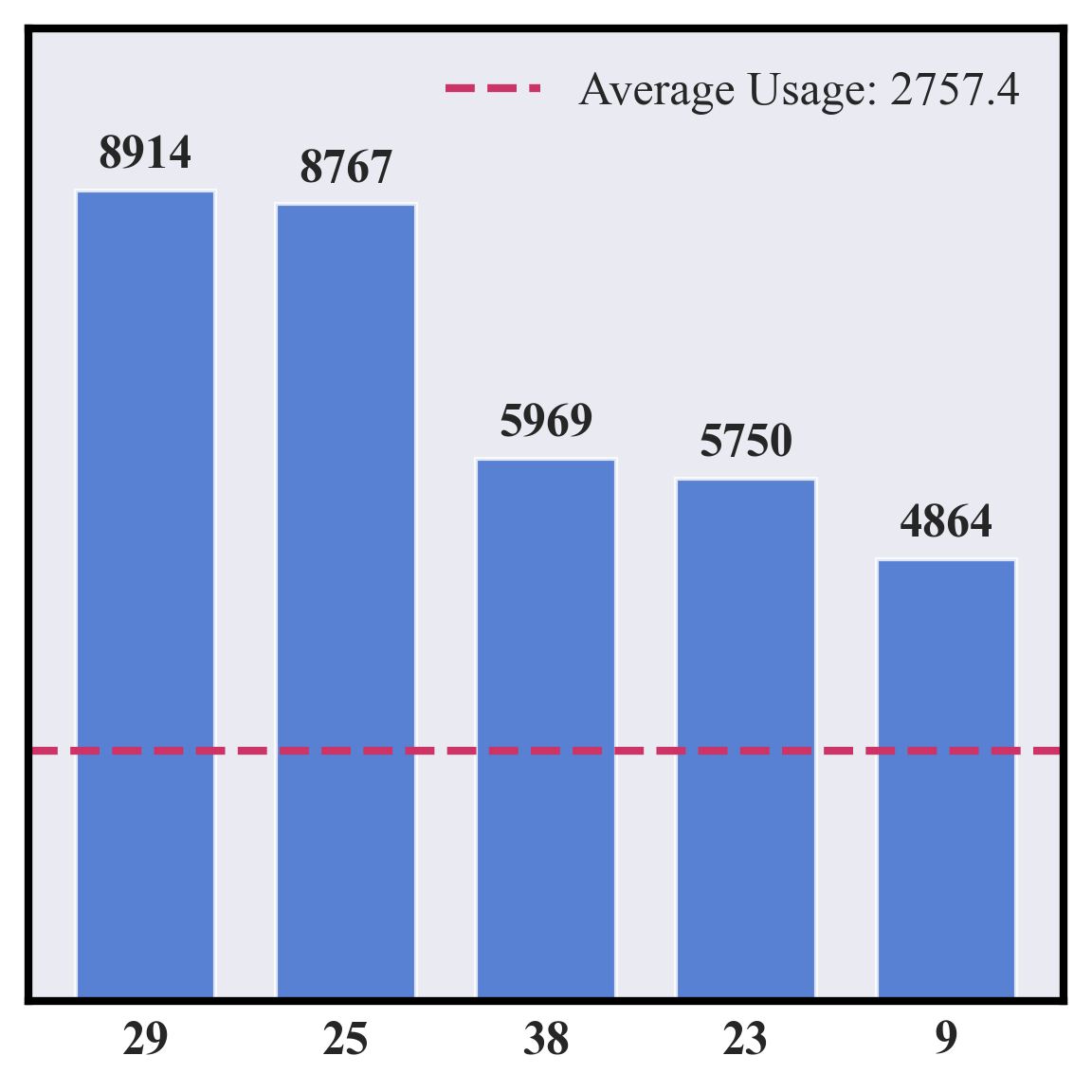}
    \caption{Domain 4}
  \end{subfigure}
    \begin{subfigure}[b]{0.19\textwidth}
    \centering
    \includegraphics[width=\linewidth]{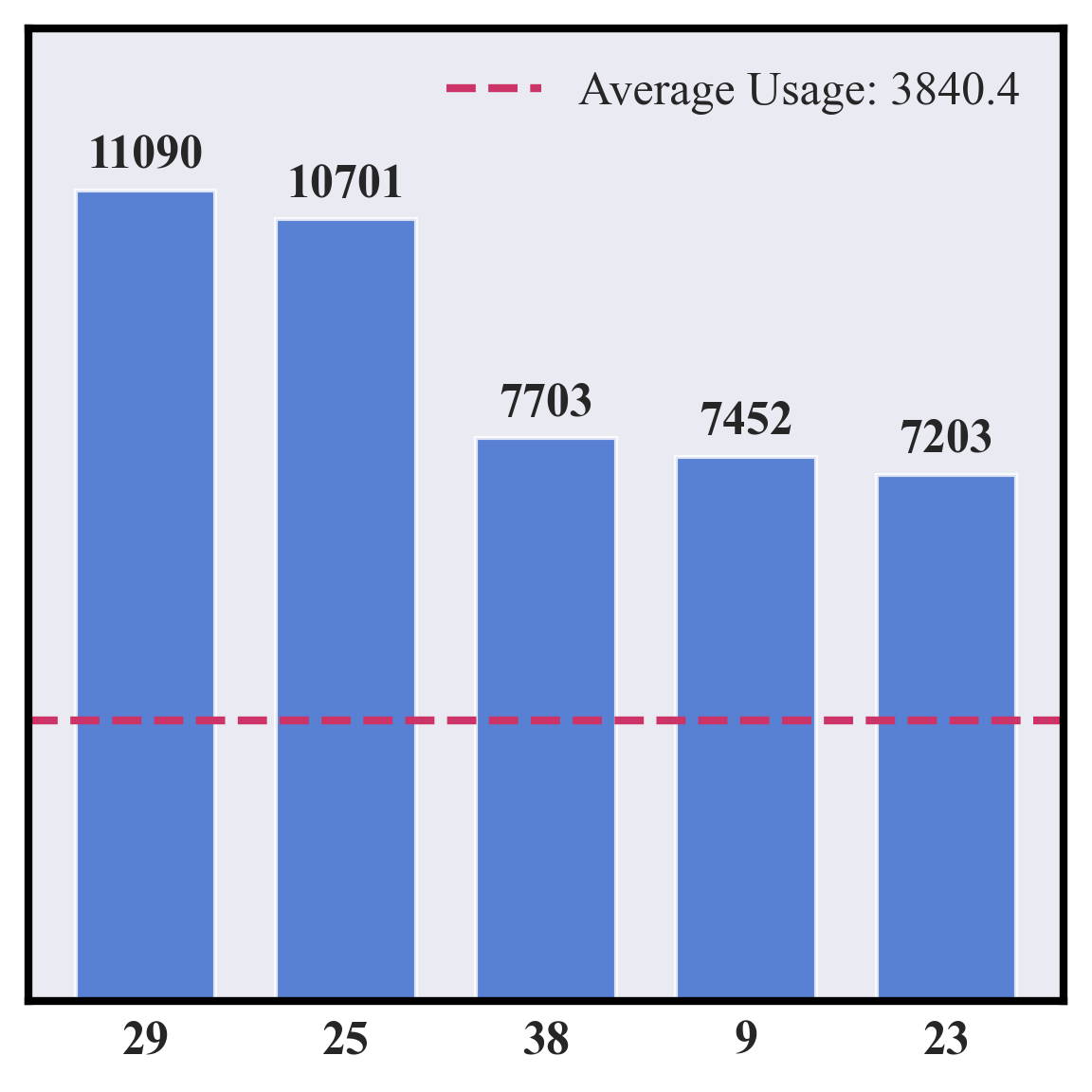}
    \caption{Domain 5}
  \end{subfigure}

  \vspace{0.6em}

  \begin{subfigure}[b]{0.19\textwidth}
    \centering
    \includegraphics[width=\linewidth]{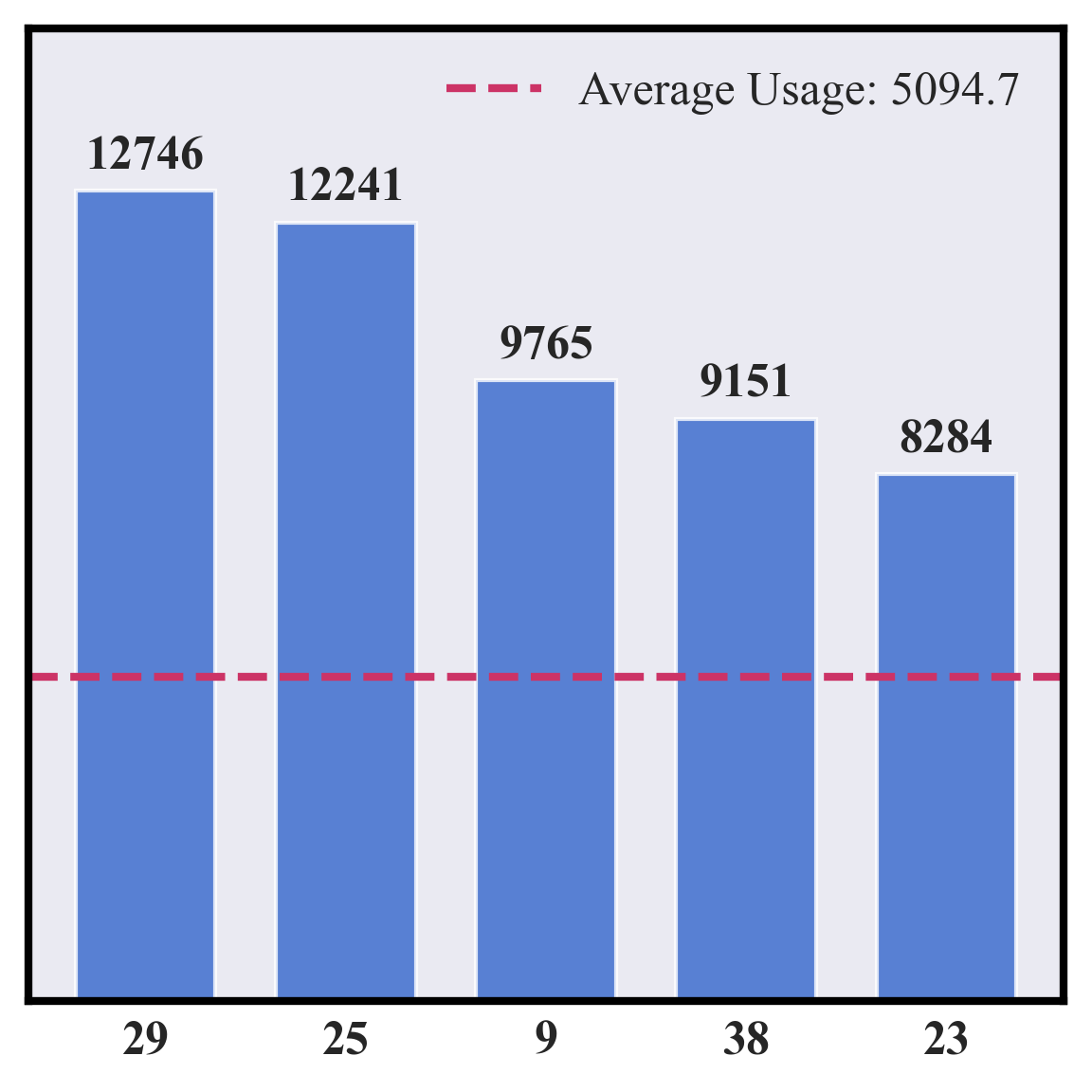}
    \caption{Domain 6}
  \end{subfigure}
  \begin{subfigure}[b]{0.19\textwidth}
    \centering
    \includegraphics[width=\linewidth]{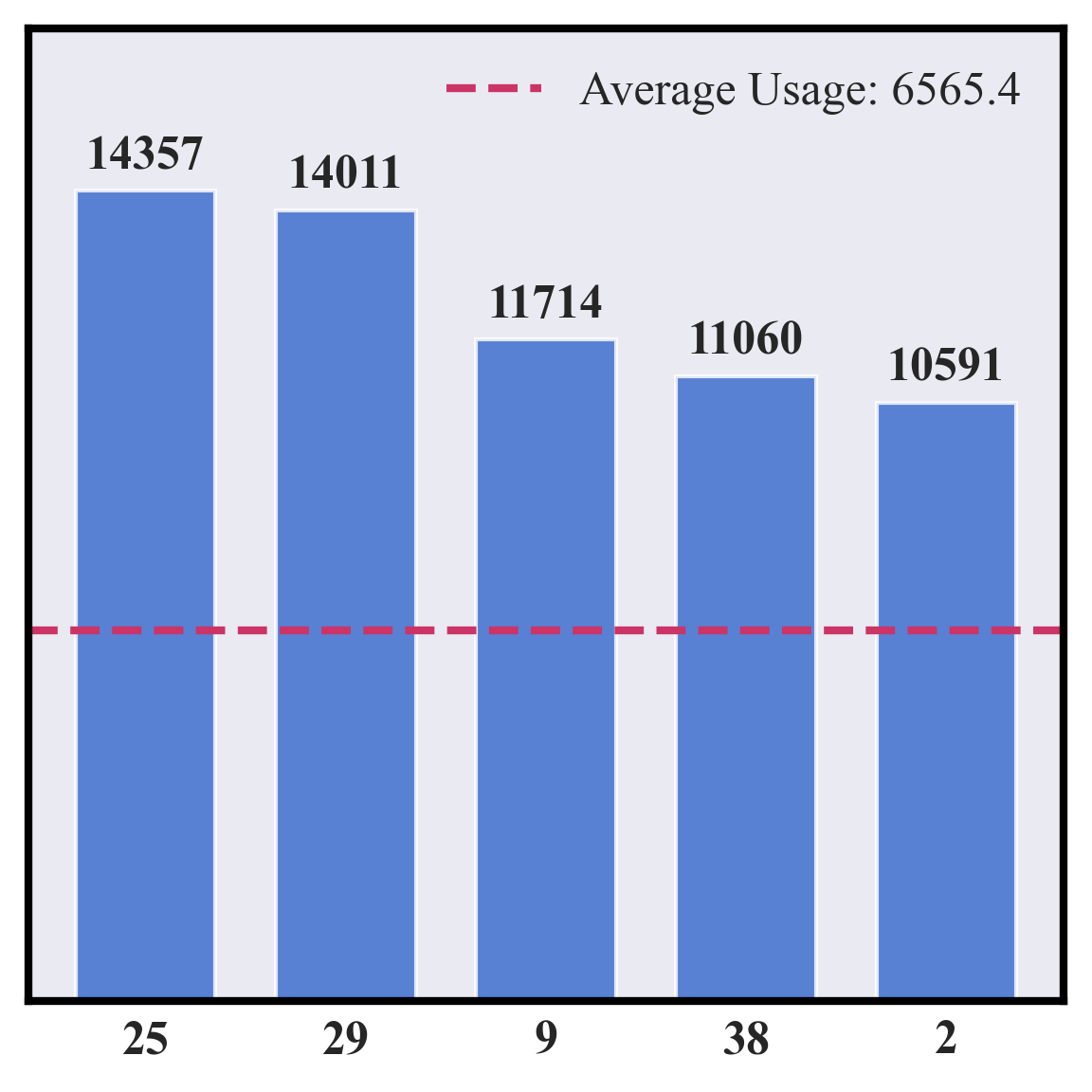}
    \caption{Domain 7}
  \end{subfigure}
  \begin{subfigure}[b]{0.19\textwidth}
    \centering
    \includegraphics[width=\linewidth]{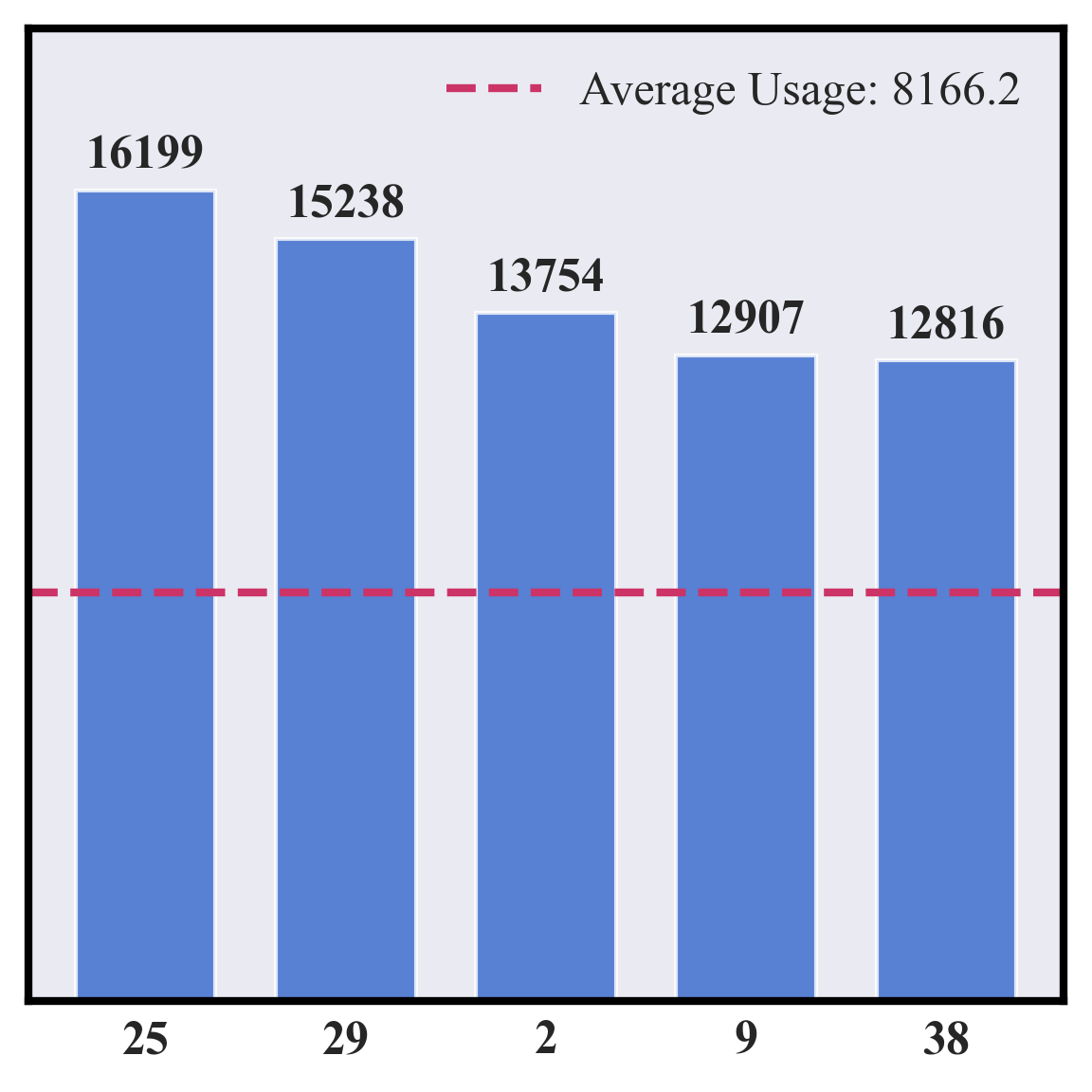}
    \caption{Domain 8}
  \end{subfigure}
  \begin{subfigure}[b]{0.19\textwidth}
    \centering
    \includegraphics[width=\linewidth]{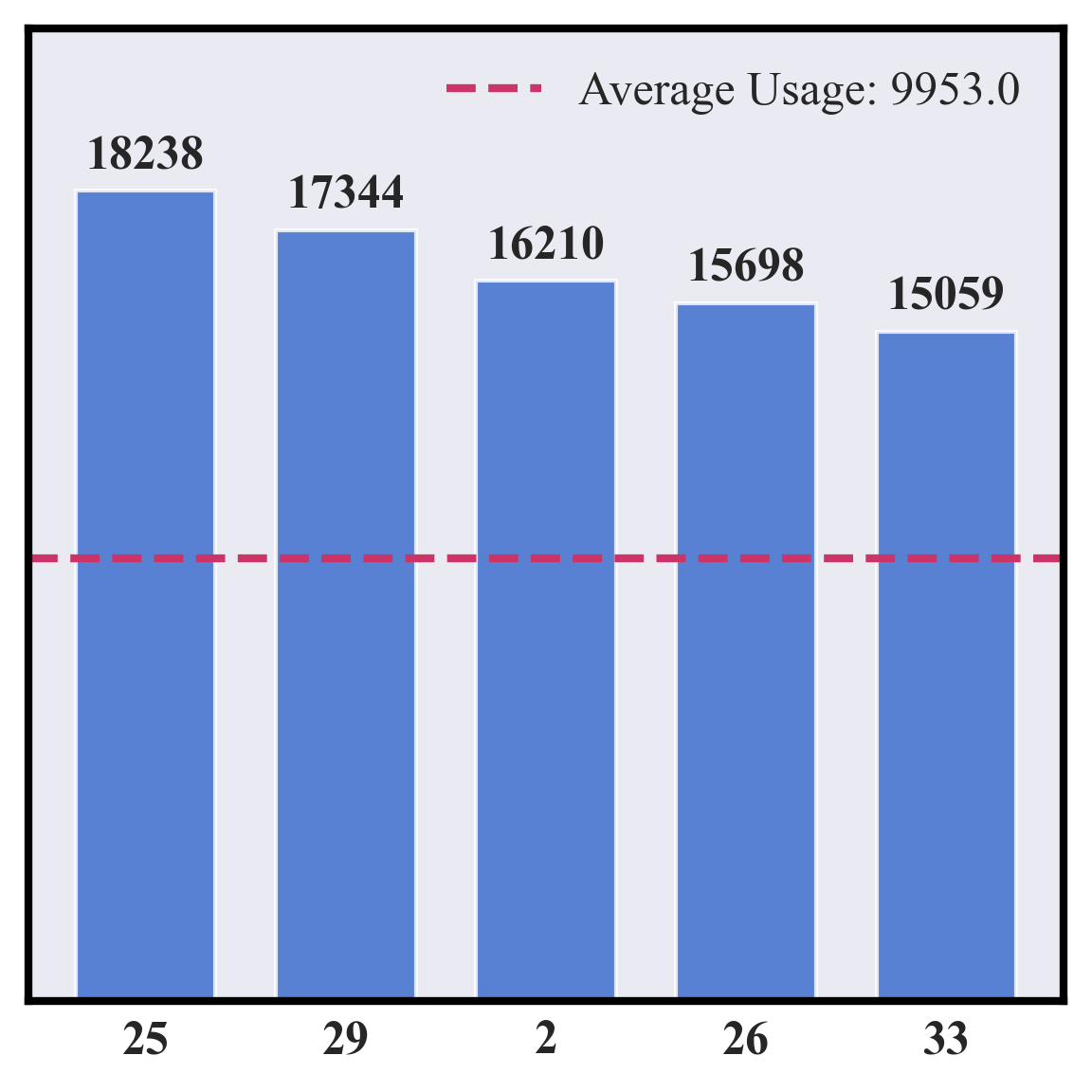}
    \caption{Domain 9}
  \end{subfigure}
    \begin{subfigure}[b]{0.19\textwidth}
    \centering
    \includegraphics[width=\linewidth]{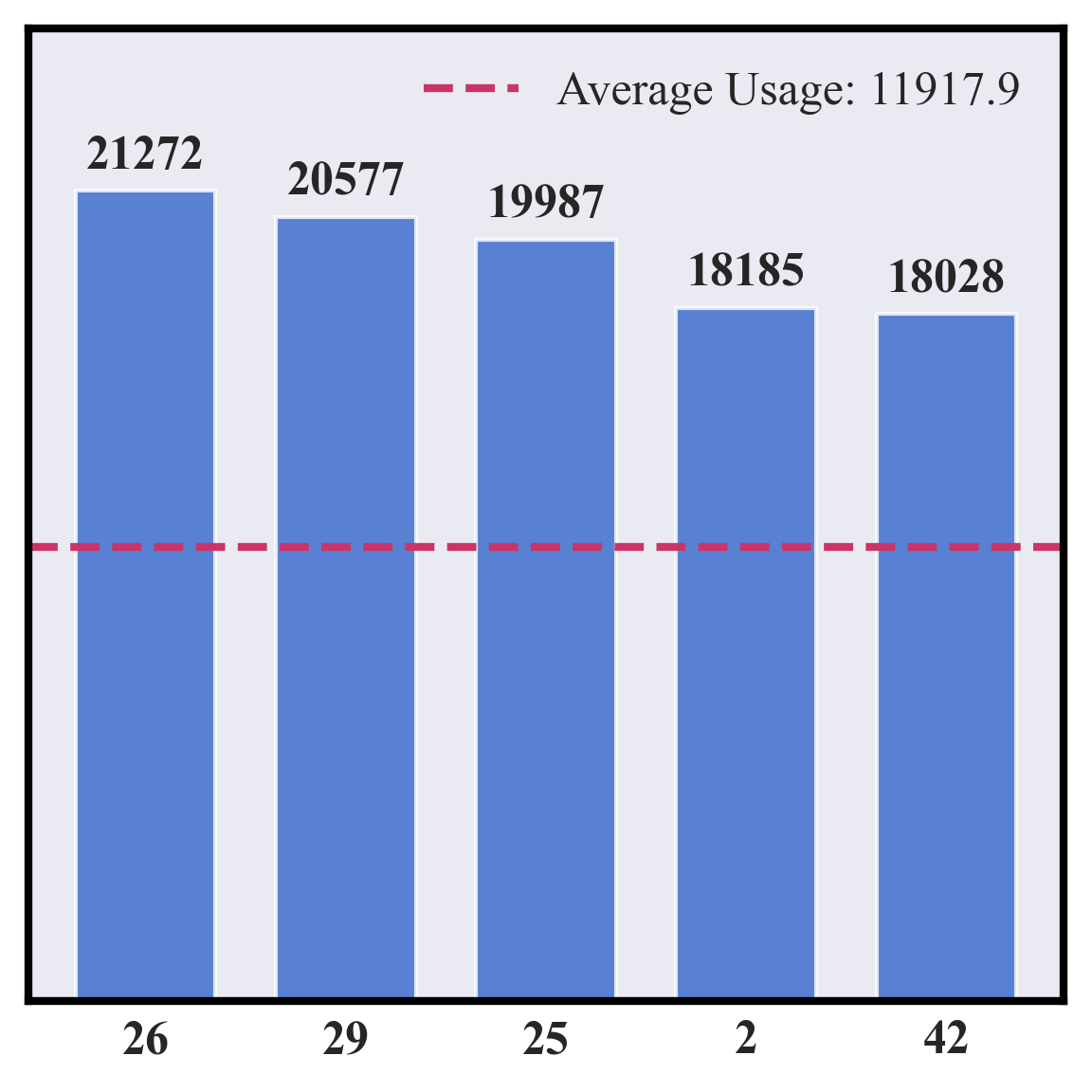}
    \caption{Domain 10}
  \end{subfigure}

  \caption{Evolution of pattern usage across Stage-Bench domains. Each subfigure displays the top-5 most frequently used evolution patterns after completing the corresponding domain.}
  \label{fig:top5p}
\end{figure*}

\begin{figure*}[!tbp]
  \centering
  \begin{subfigure}[b]{0.49\textwidth}
    \centering
    \includegraphics[width=\linewidth]{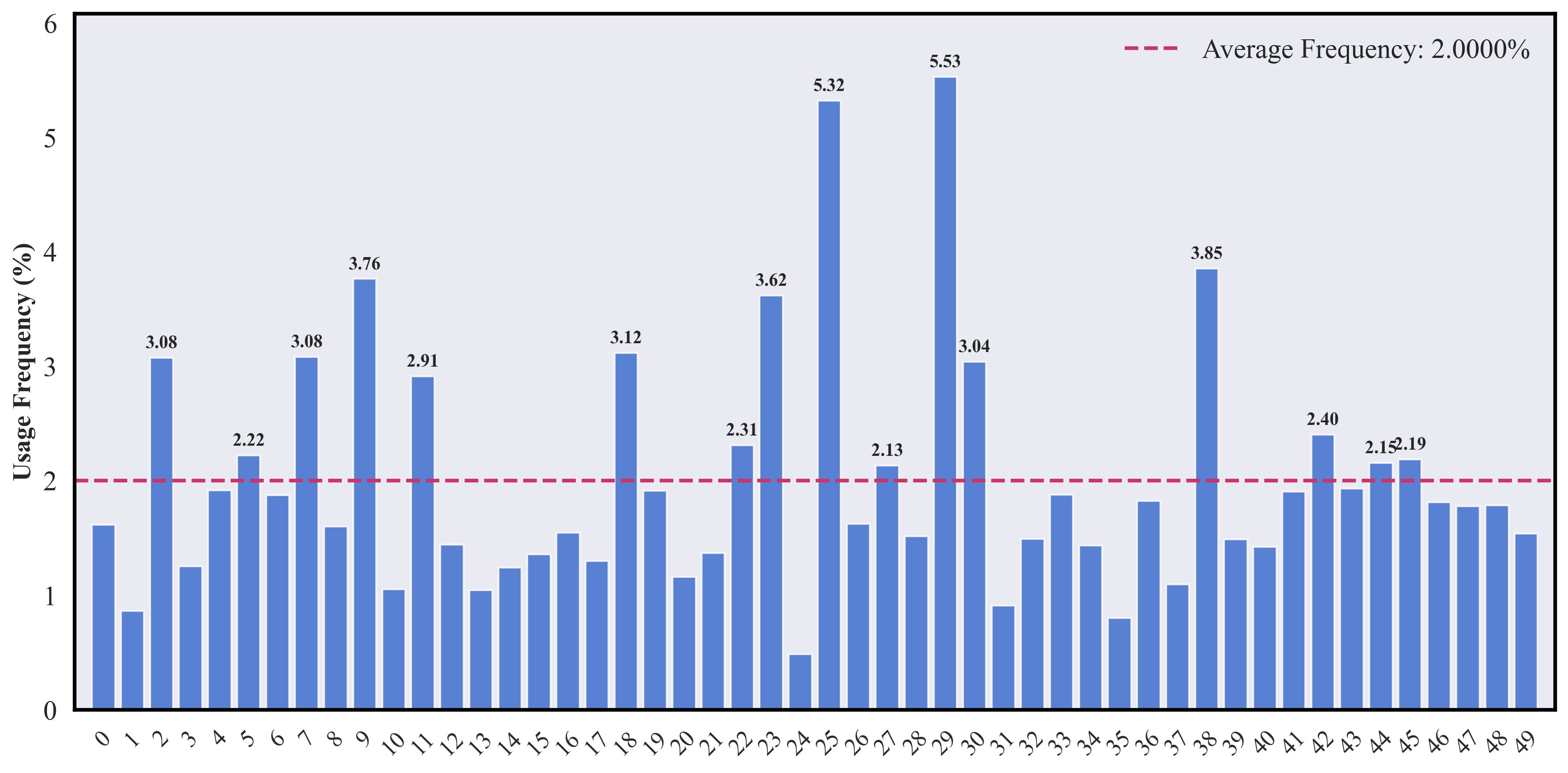}
    \caption{After Domain 5}
  \end{subfigure}
  \begin{subfigure}[b]{0.49\textwidth}
    \centering
    \includegraphics[width=\linewidth]{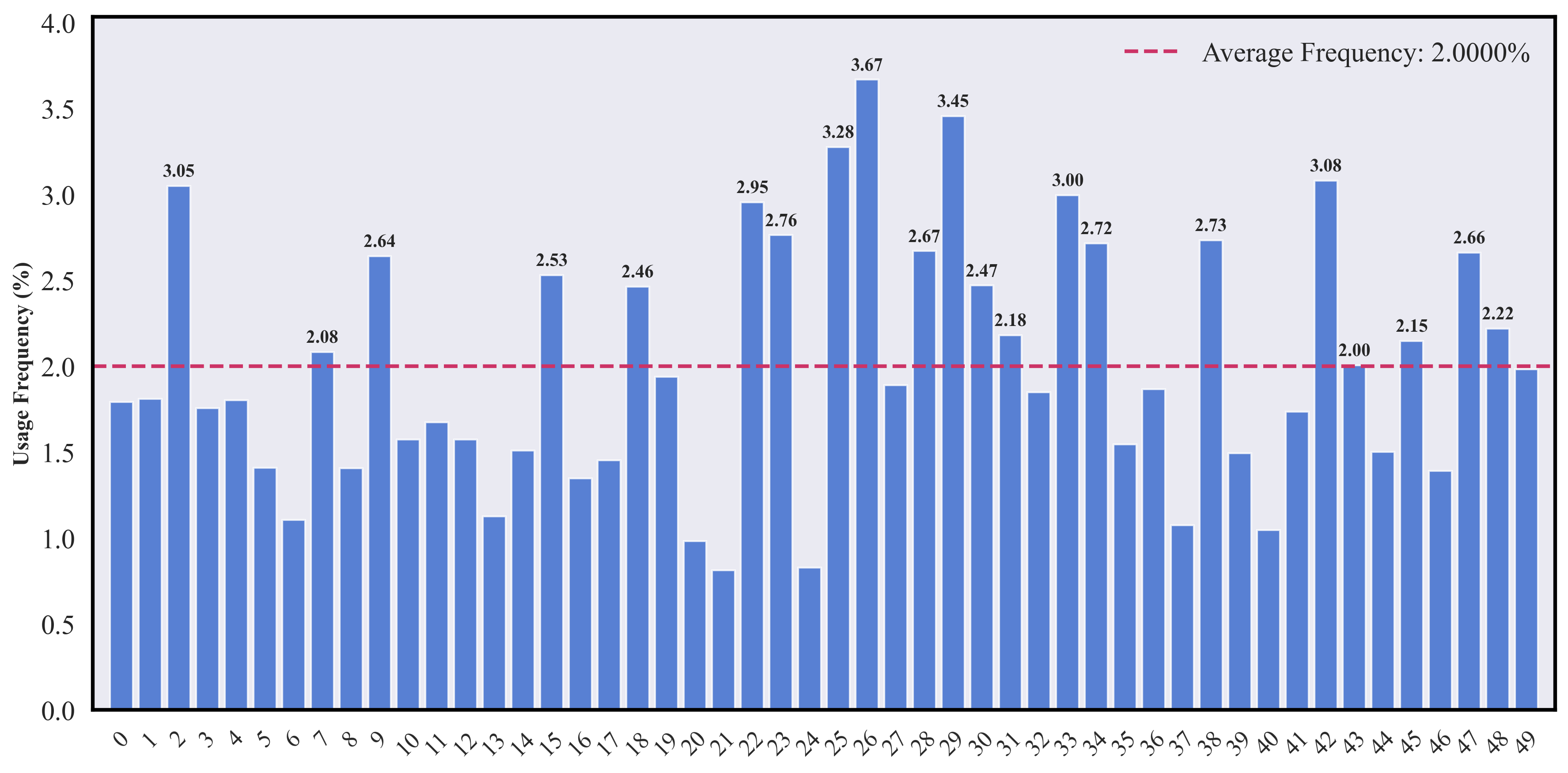}
    \caption{After Domain 10}
  \end{subfigure}
    \caption{Evolution of the pattern-usage distribution. Usage is concentrated after Domain 5 (left) and becomes more balanced after Domain 10 (right).}
  \label{fig:total_pattern}
\end{figure*}

\subsection{Pattern-Pool Dynamics}
For completeness, after the gradient update produces an intermediate selected pattern $\widetilde r_i$, STAGE applies the following no-gradient decay-and-injection update:
\begin{equation}
  r_i \leftarrow (1-\eta)\widetilde r_i
  + \eta w_i^y\operatorname{stopgrad}(\Delta^y),
  \qquad i\in\operatorname{top}\text{-}k,
  \label{eq:pattern_update}
\end{equation}
where $\eta$ is the update rate, $w_i^y$ is the attention weight of pattern $r_i$ for class $y$, and $\Delta^y$ is the observed class-level evolution displacement. Patterns outside the selected set are unchanged by this online update. The shared retention coefficient distinguishes Eq.~\eqref{eq:pattern_update} from a standard attention-weighted EMA.

To examine how STAGE uses its evolution patterns over time, we record how often each pattern enters the \textbf{top-$k$} set and track the resulting frequency distribution during training. Fig.~\ref{fig:top5p} and Fig.~\ref{fig:total_pattern} reveal an empirical progression that we summarize as \emph{uniform} $\to$ \emph{differentiation} $\to$ \emph{re-balancing}.

During \emph{cold start}, pattern parameters remain close to their random initialization. Different patterns consequently have similar cosine similarities to the class anchors and comparable probabilities of entering the top-$k$ set. Usage variance is low because no pattern has yet specialized to a particular morphological change.

During \emph{differentiation}, stochastic retrieval and early task exposure cause a subset of patterns to be selected more often. Selected patterns receive gradients through the attention context and the non-gradient decay-and-injection update in Eq.~\eqref{eq:pattern_update}. Their closer alignment with frequently observed evolution vectors increases the chance of subsequent selection, producing a ``rich-get-richer'' effect. The resulting rise in usage variance indicates that a small subset of patterns is specializing to reusable evolution directions.

During \emph{re-balancing}, new classes and domains introduce displacements that need not align with previously dominant patterns. The competitive update decays every selected pattern by the same retention factor while injecting the new displacement in proportion to its attention weight. A previously dominant but weakly matched pattern therefore receives less reinforcement, allowing other patterns to enter the top-$k$ set and specialize. The stored class anchors remain fixed throughout this process.

Overall, the observed dynamics show that the pool does not collapse to a single dominant direction in this experiment. Its fixed capacity is redistributed over the stream under the same hard top-$k$ retrieval rule used in the main paper.

\subsection{Computational and Memory Complexity}
We next analyze the computational and memory overhead introduced by STAGE on top of the frozen CLIP backbone. Fig.~\ref{fig:time} reports end-to-end training time under the shared ViT-B/16 backbone and protocol. STAGE incurs moderate overhead relative to the PTM-based CIL baselines while remaining within the same order of magnitude.

Formally, let $d$ denote the feature dimension, $C$ the number of classes seen so far, $K$ the total number of patterns in the shared pool ($K{=}50$ in the main setup), and $k$ the number selected by top-$k$ retrieval ($k{=}5$ by default). Scoring an anchor against all patterns costs $\mathcal{O}(K d)$. Applying the three-layer evolution MLP of hidden width $h$ to the selected context costs approximately $\mathcal{O}(k d h)$. The per-sample cost attributable to retrieval and evolution prediction is therefore
\begin{equation}
\mathcal{O}(K d + k d h),
\end{equation}
where $K$, $k$, and $h$ are fixed in our experiments. This expression isolates the evolution module; the measured end-to-end cost, including projection and fusion operations, is reported in Fig.~\ref{fig:time}.

The data-dependent memory of STAGE stores compact class anchors and one shared pattern pool rather than raw training images. Anchors require $\mathcal{O}(C d)$ memory, while the pattern pool requires $\mathcal{O}(K d)$ memory and is constant with respect to $C$. The resulting data-dependent storage is
\begin{equation}
\mathcal{O}(C d + K d),
\end{equation}
This accounting excludes the fixed-size parameters of the auxiliary networks and the task-specific projection heads, which are model parameters rather than stored training examples. It makes explicit that the evolution-pattern memory itself does not grow with the number of classes.

\section{Additional Experimental Results}

\subsection{Stage-Aware Diagnostic Controls}
To separate the effect of stage metadata from that of explicit transition modeling, we augment PROOF with two lightweight controls under the same backbone and protocol. \emph{Stage Cue} exposes the stage indicator during training, while \emph{Stage Prototype} additionally maintains separate prototypes for the observed stages. At test time, both controls remain stage-agnostic and predict the semantic class without a ground-truth stage label. Table~\ref{tab:stage_controls} shows that both controls improve PROOF but remain substantially below STAGE. In this diagnostic, stage metadata and static stage-specific prototypes are therefore insufficient to explain the gain obtained by explicit transition modeling.

\begin{table}[t]
  \centering
  \caption{Stage-aware diagnostic controls on Stage-Bench.}
  \label{tab:stage_controls}
  \setlength{\tabcolsep}{5pt}
  \begin{tabular}{lcc}
    \toprule
    Method & Final Acc. (\%) $\uparrow$ & Intra-F (\%) $\downarrow$ \\
    \midrule
    PROOF & 37.52 & 29.37 \\
    PROOF + Stage Cue & 39.45 & 25.82 \\
    PROOF + Stage Prototype & 41.12 & 22.45 \\
    \textbf{STAGE} & \textbf{56.44} & \textbf{7.48} \\
    \bottomrule
  \end{tabular}
\end{table}

\subsection{Class-Order Robustness}
Because continual-learning results can depend on the order of incoming classes, we repeat the Stage-Bench evaluation over five class-order seeds. As reported in Table~\ref{tab:class_order}, STAGE retains a clear advantage over MOS in both final accuracy and Intra-F, while exhibiting comparable or lower variability.

\begin{table}[t]
  \centering
  \caption{Mean and standard deviation over five class-order seeds.}
  \label{tab:class_order}
  \setlength{\tabcolsep}{6pt}
  \begin{tabular}{lcc}
    \toprule
    Method & Final Acc. (\%) $\uparrow$ & Intra-F (\%) $\downarrow$ \\
    \midrule
    MOS & $41.23{\pm}5.44$ & $24.20{\pm}3.59$ \\
    \textbf{STAGE} & $\mathbf{56.71{\pm}4.36}$ & $\mathbf{7.48{\pm}0.94}$ \\
    \bottomrule
  \end{tabular}
\end{table}

\begin{figure}[!tbp]
  \centering
  \begin{subfigure}{\columnwidth}
    \centering
    \includegraphics[width=\linewidth]{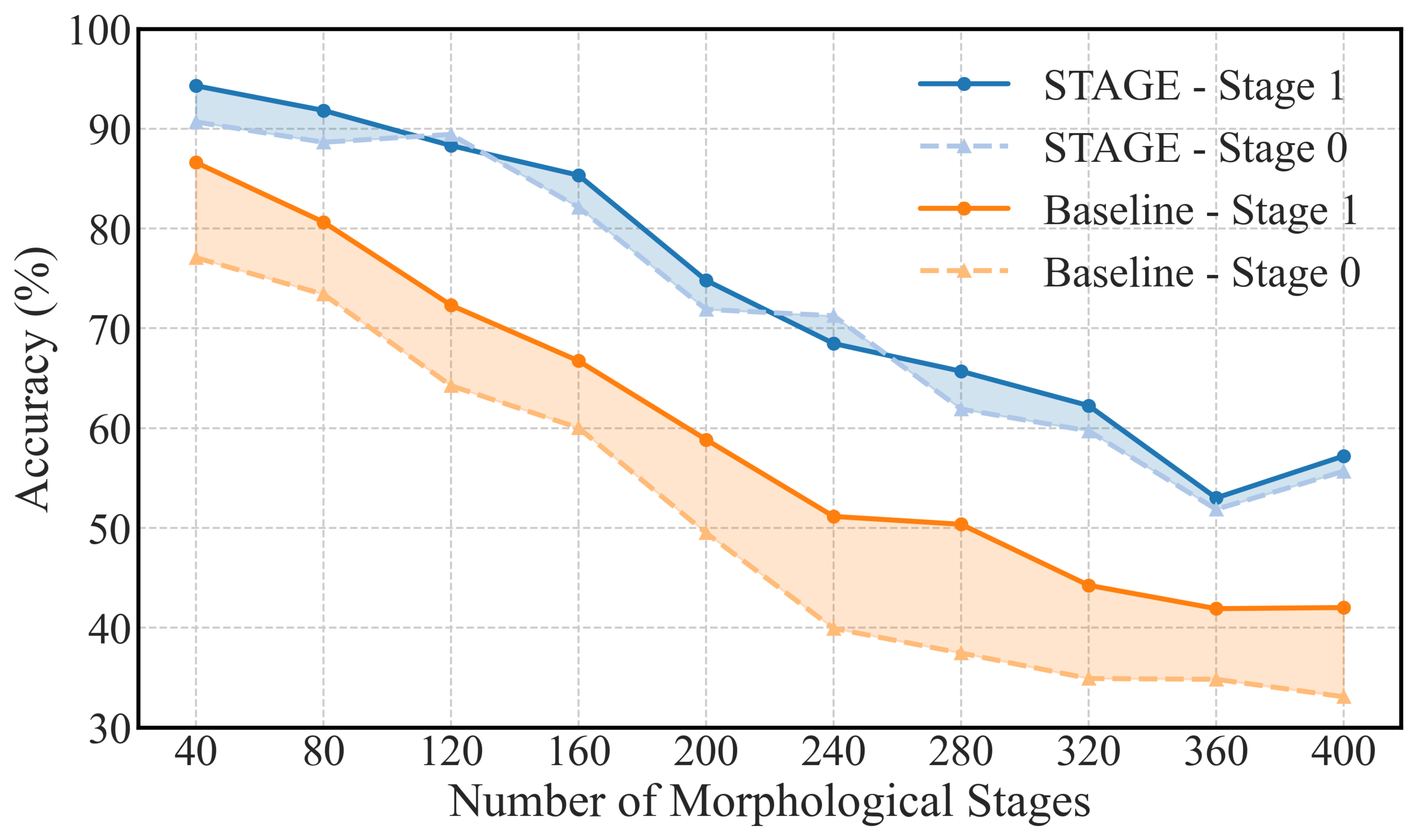}
    \caption{Stage-wise accuracy throughout the incremental stream.}
    \label{fig:supp_intra}
  \end{subfigure}
  \vspace{0.8em}

  \begin{subfigure}{\columnwidth}
    \centering
    \includegraphics[width=\linewidth]{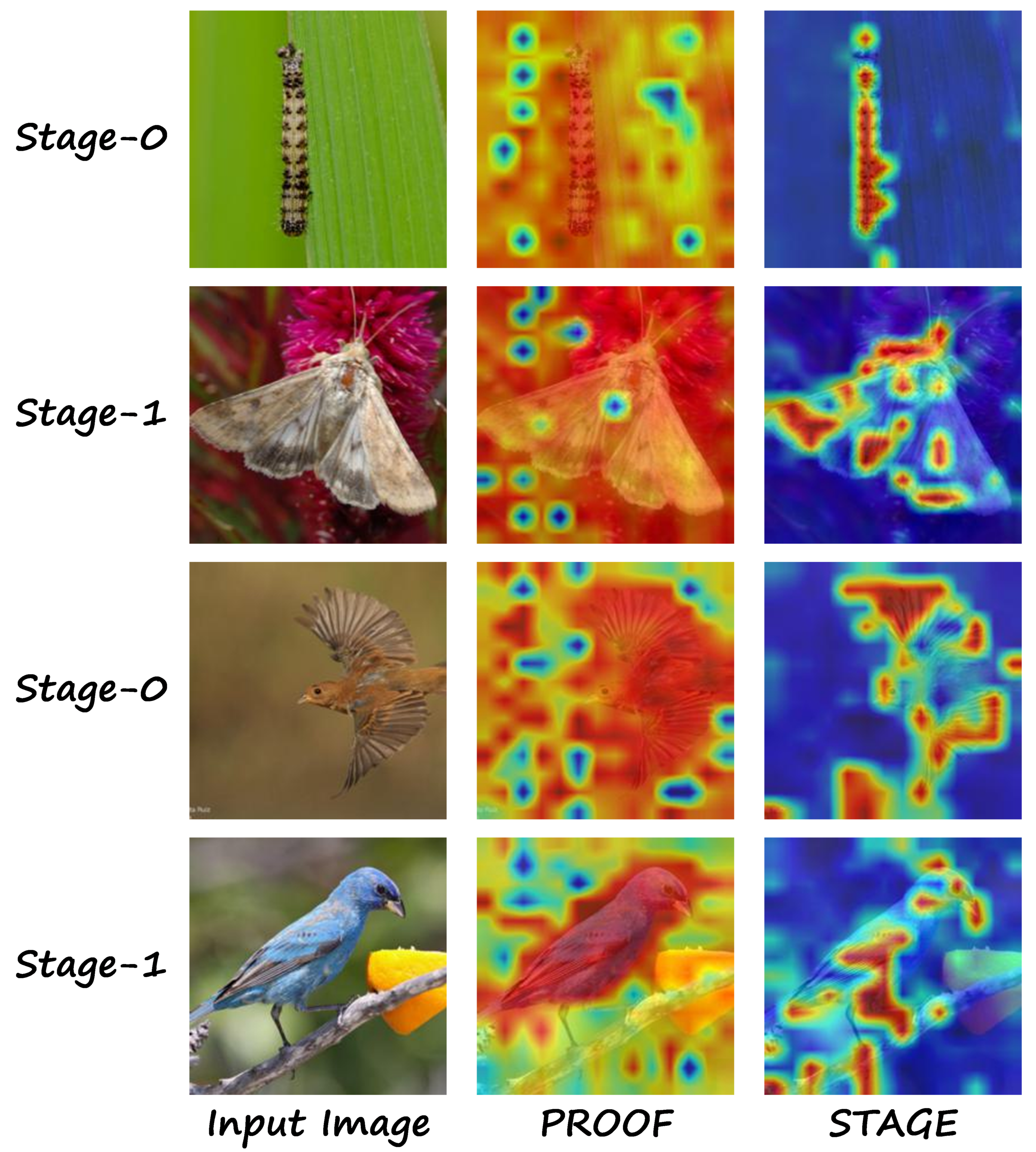}
    \caption{Grad-CAM comparison between PROOF and STAGE.}
    \label{fig:supp_attention}
  \end{subfigure}
  \caption{Qualitative analysis of intra-class forgetting and stage-dependent attention.}
  \label{fig:supp_visualization}
\end{figure}

\begin{figure*}[!tbp]
  \centering
  \begin{subfigure}[b]{0.245\textwidth}
    \centering
    \includegraphics[width=\linewidth]{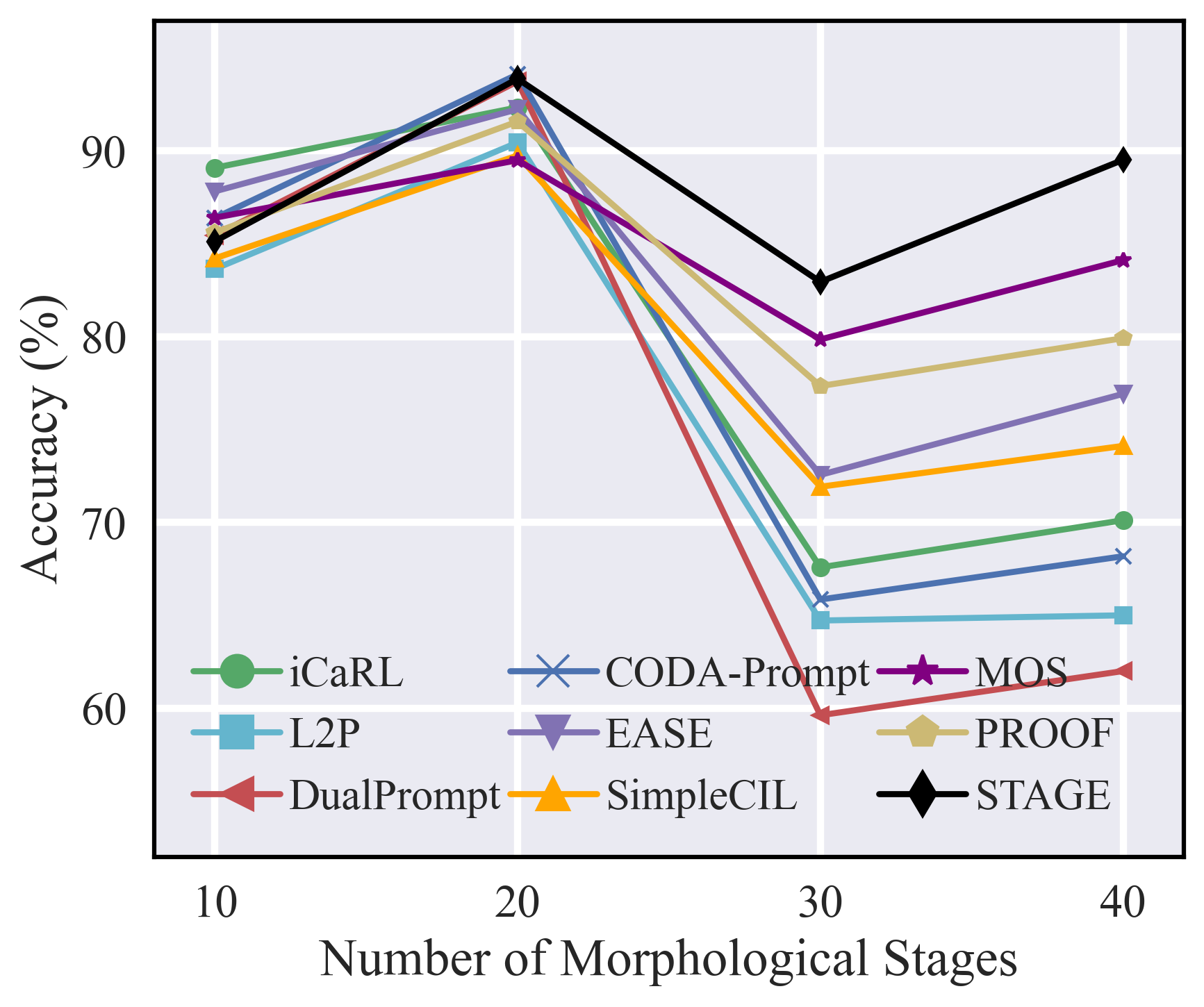}
    \caption{Fish}
  \end{subfigure}
  \begin{subfigure}[b]{0.245\textwidth}
    \centering
    \includegraphics[width=\linewidth]{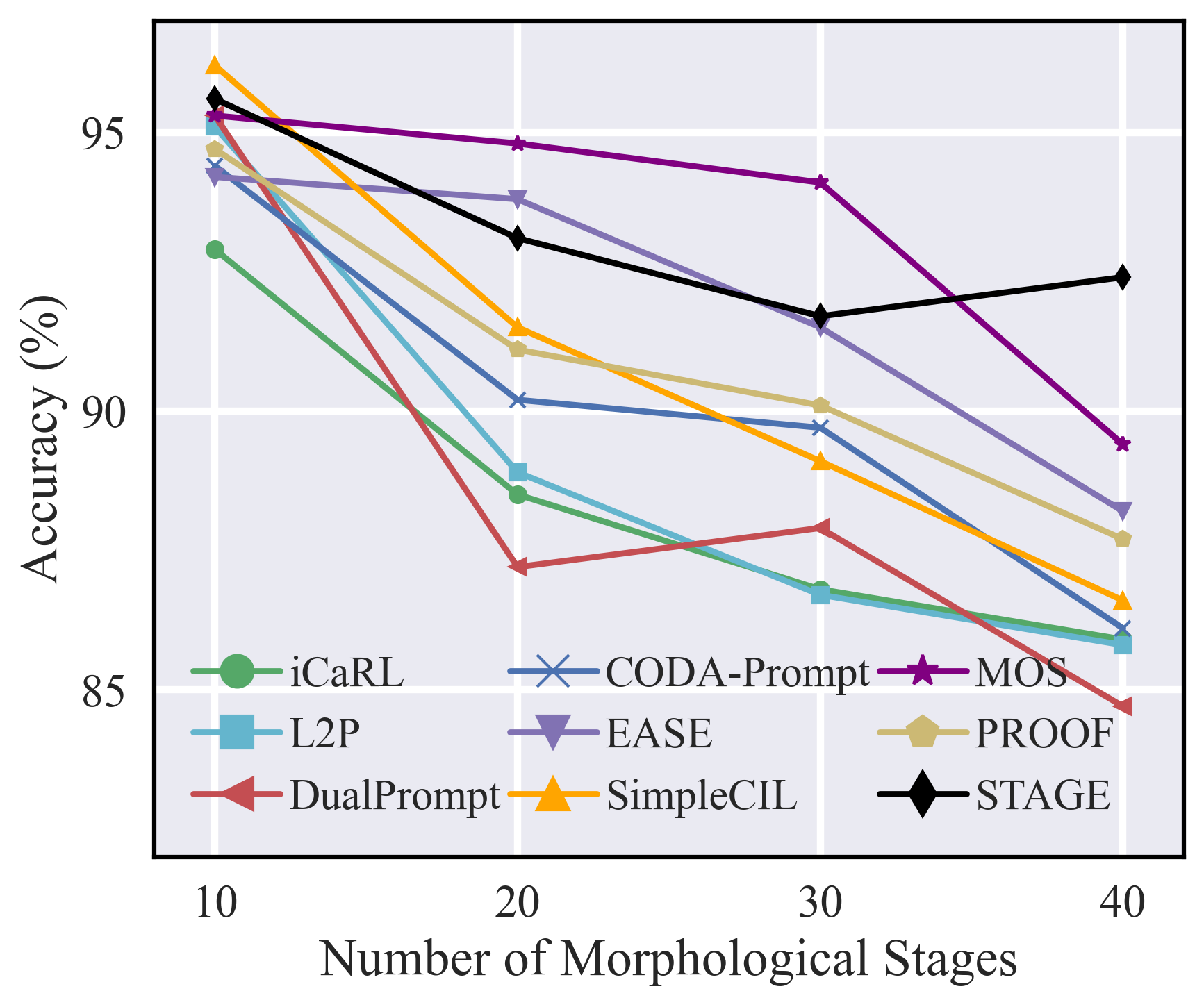}
    \caption{Flower}
  \end{subfigure}
  \begin{subfigure}[b]{0.245\textwidth}
    \centering
    \includegraphics[width=\linewidth]{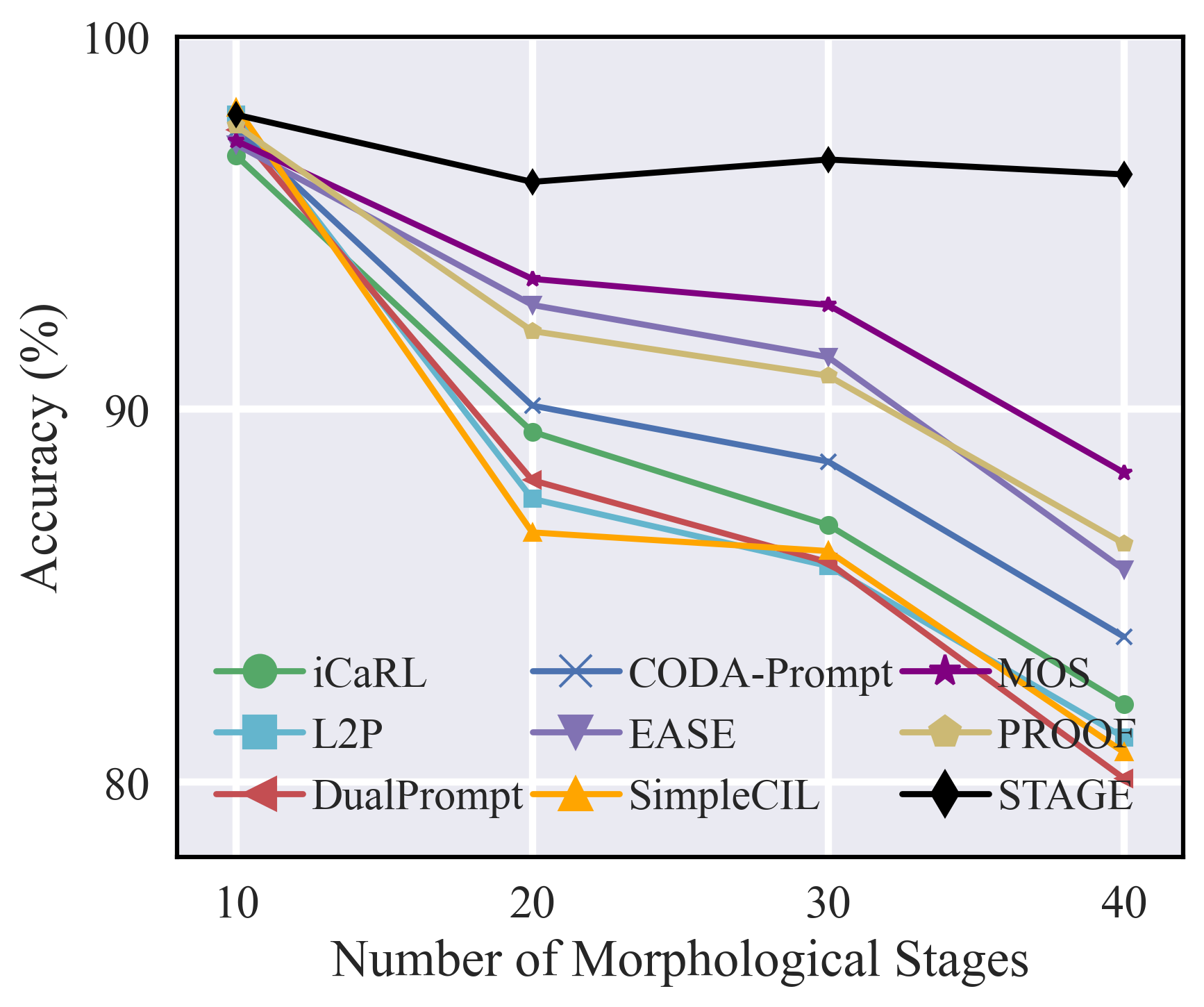}
    \caption{Food}
  \end{subfigure}
  \begin{subfigure}[b]{0.245\textwidth}
    \centering
    \includegraphics[width=\linewidth]{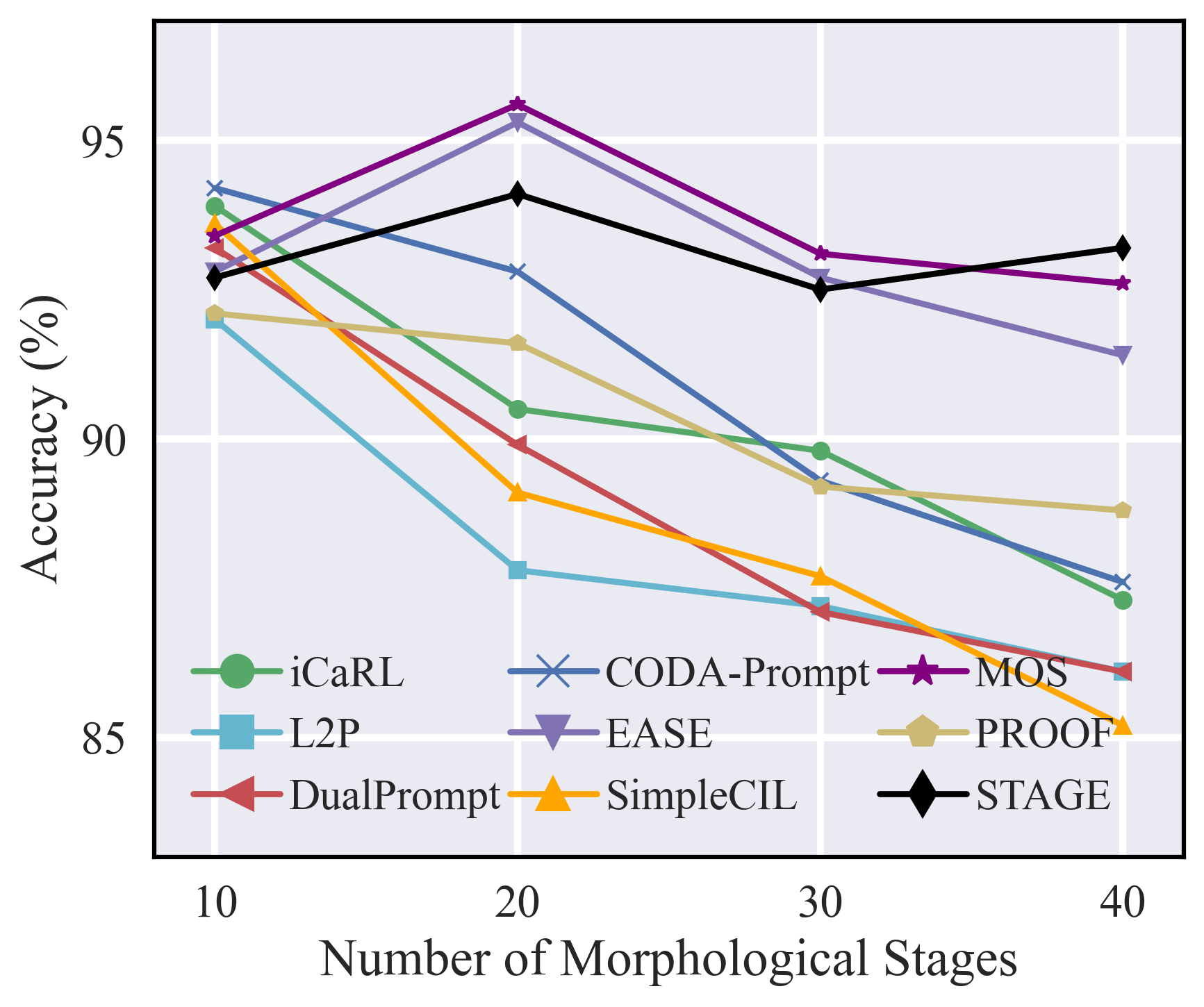}
    \caption{VegFru}
  \end{subfigure}

  \vspace{0.6em}

  \begin{subfigure}[b]{0.245\textwidth}
    \centering
    \includegraphics[width=\linewidth]{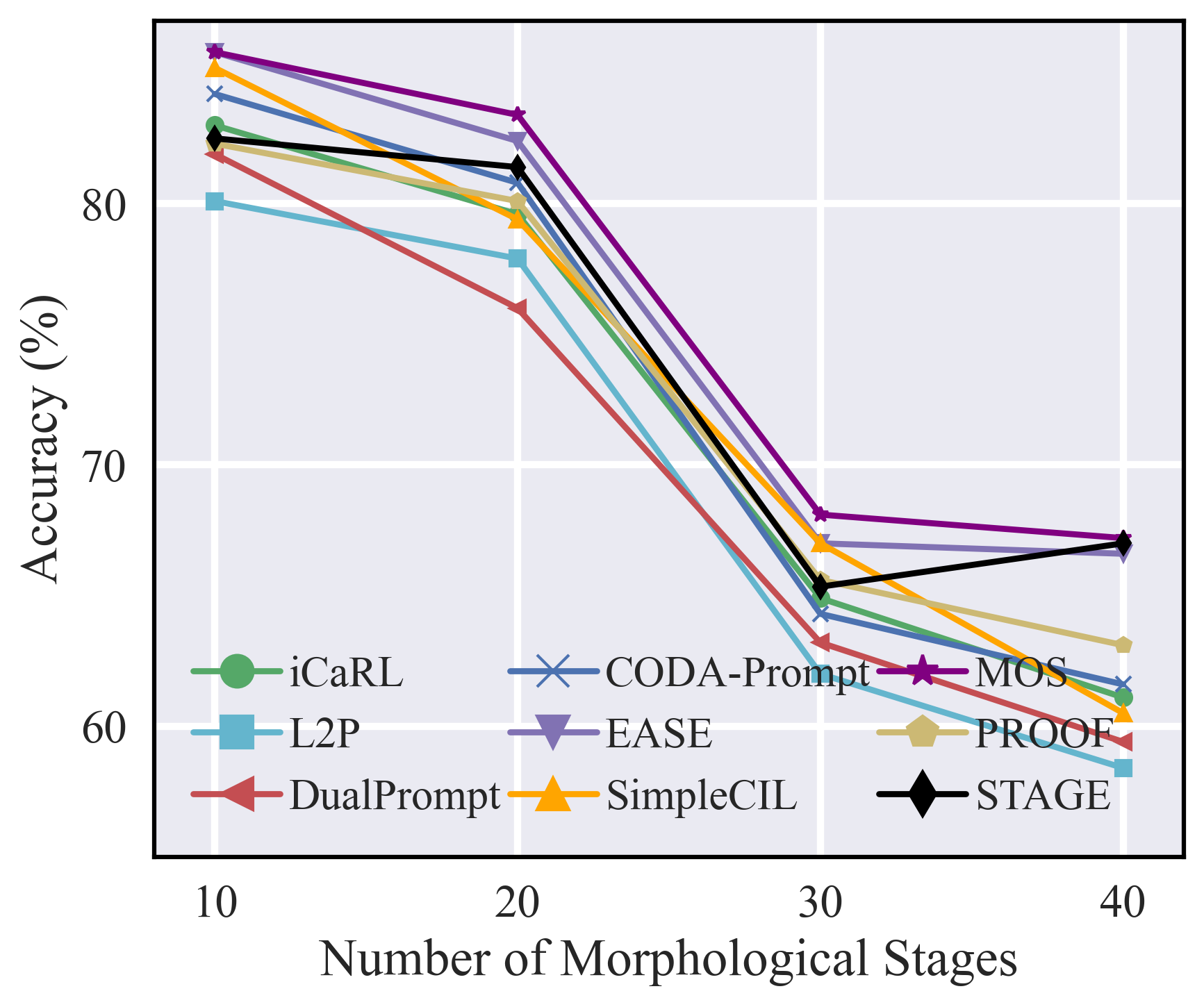}
    \caption{Fungi}
  \end{subfigure}
  \begin{subfigure}[b]{0.245\textwidth}
    \centering
    \includegraphics[width=\linewidth]{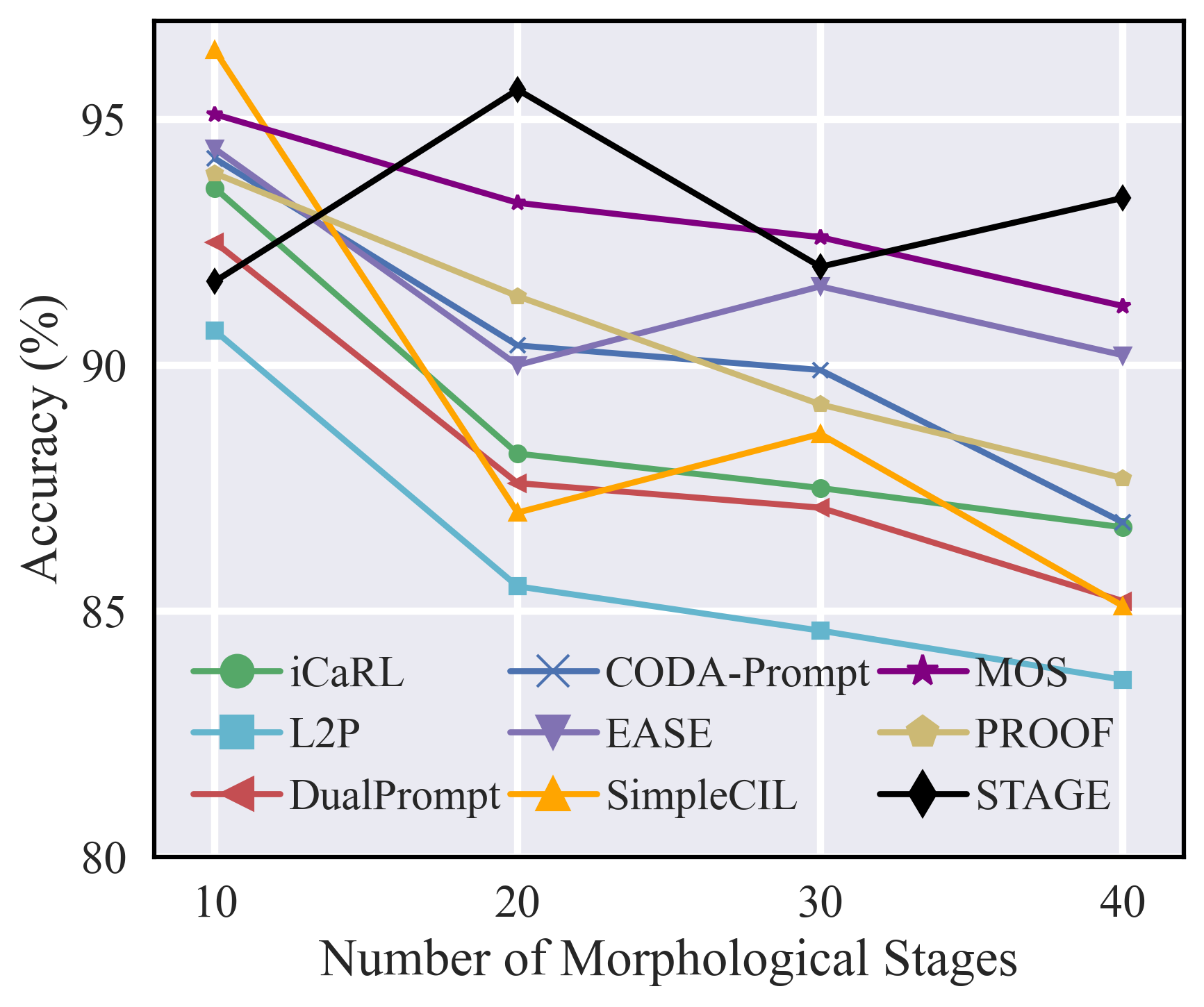}
    \caption{Birds}
  \end{subfigure}
  \begin{subfigure}[b]{0.245\textwidth}
    \centering
    \includegraphics[width=\linewidth]{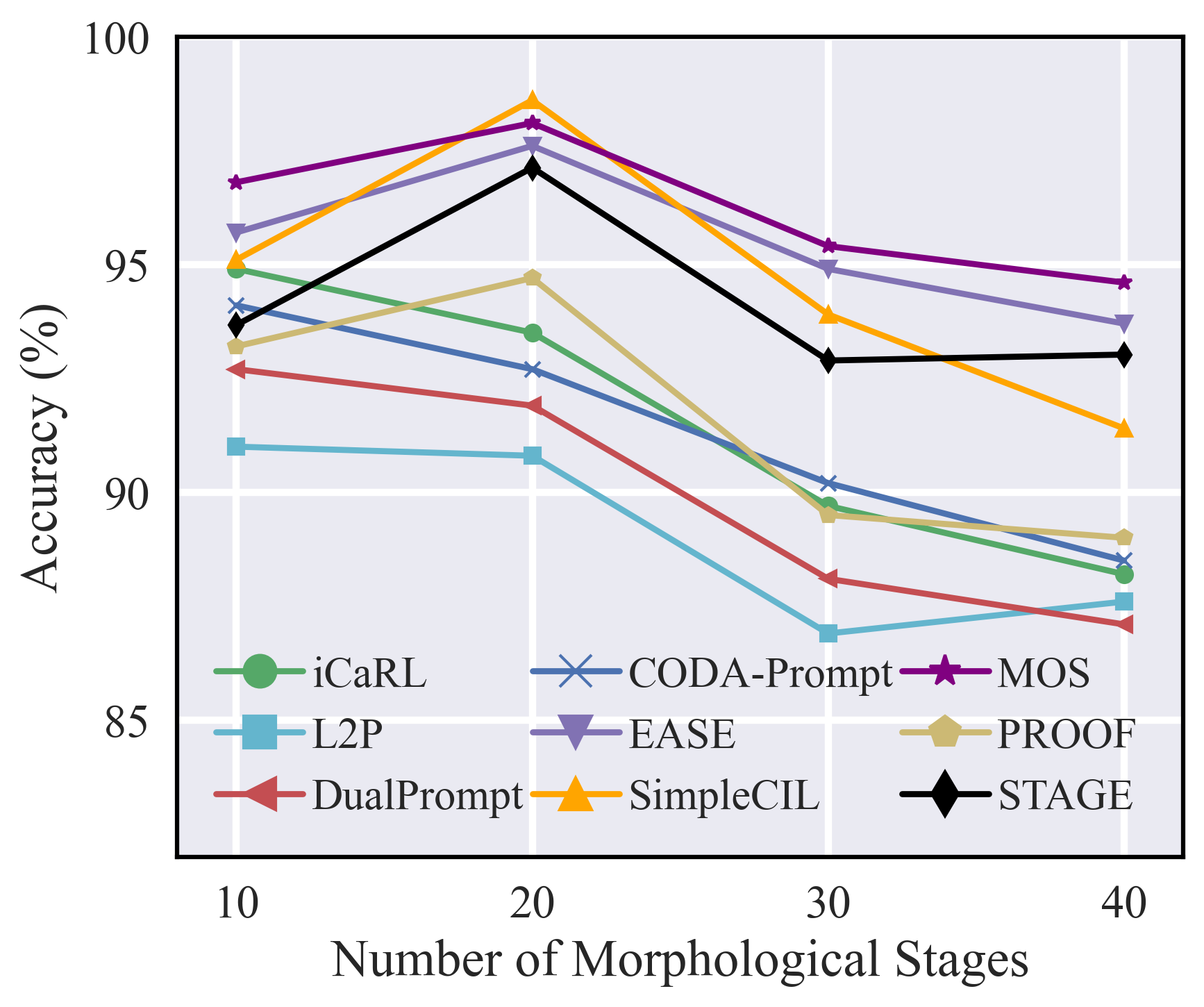}
    \caption{Pets}
  \end{subfigure}
  \begin{subfigure}[b]{0.245\textwidth}
    \centering
    \includegraphics[width=\linewidth]{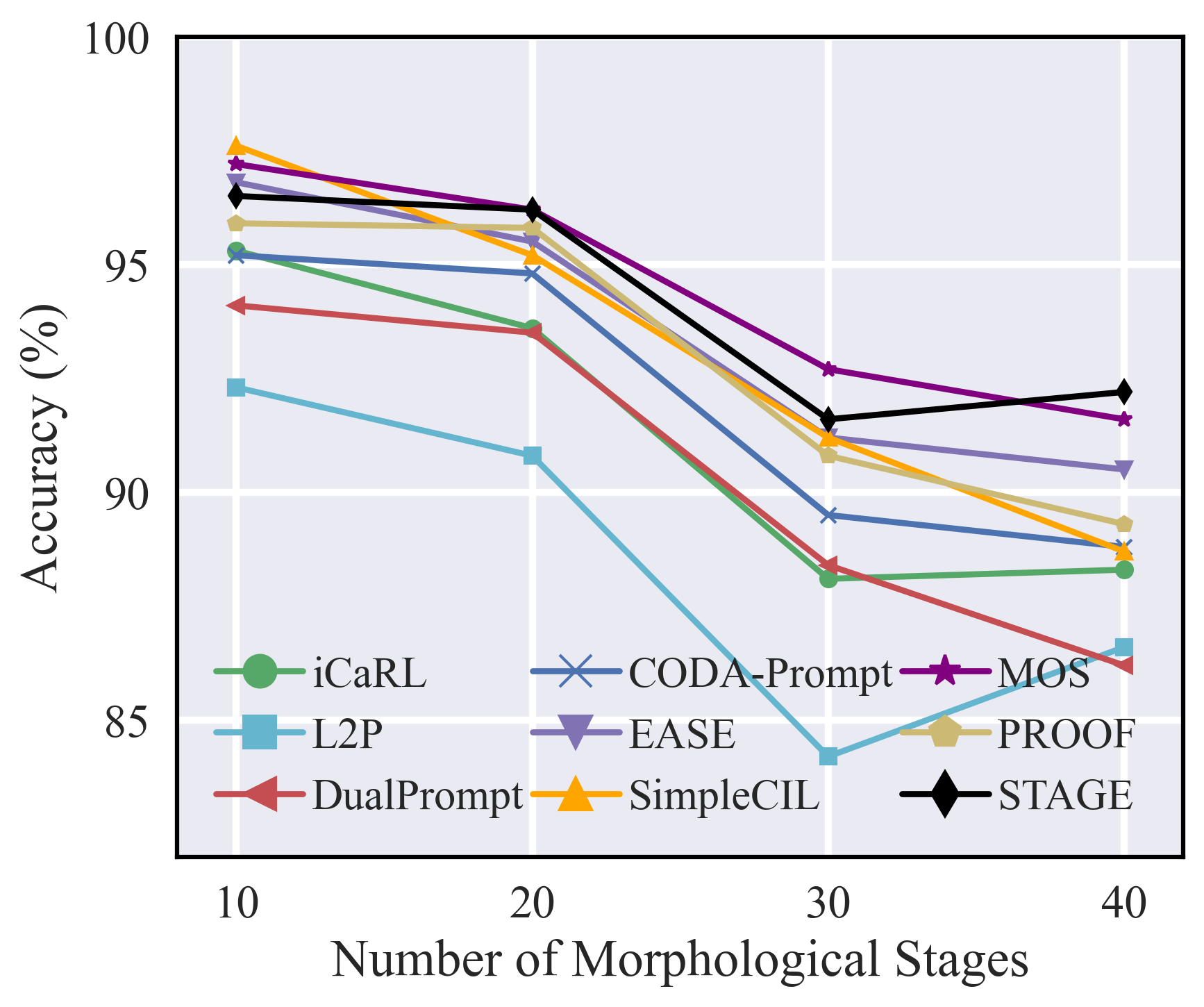}
    \caption{PlantLeaves}
  \end{subfigure}

  \vspace{0.6em}
  \begin{subfigure}[b]{0.245\textwidth}
    \centering
    \includegraphics[width=\linewidth]{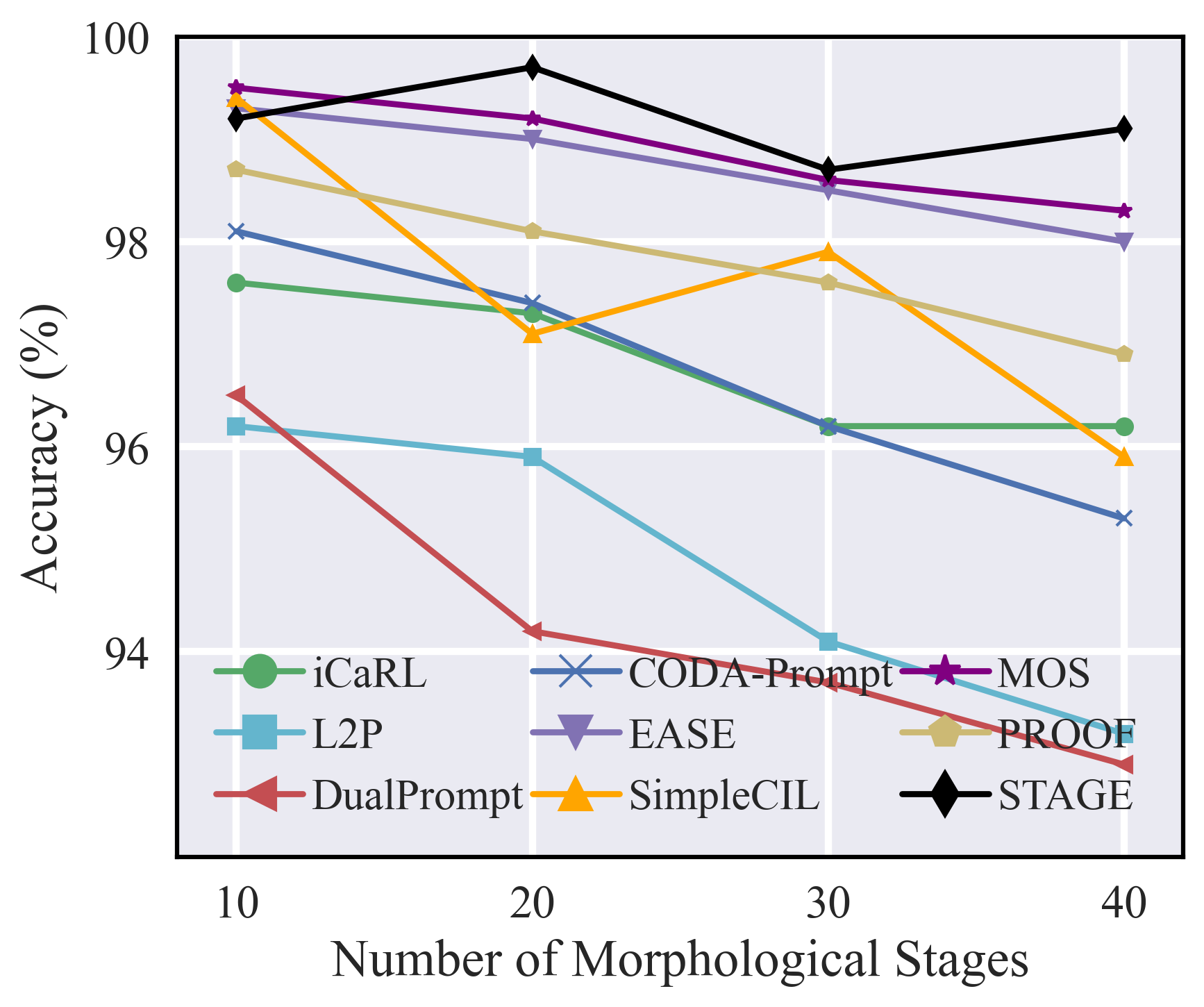}
    \caption{Objects}
  \end{subfigure}
  \begin{subfigure}[b]{0.245\textwidth}
    \centering
    \includegraphics[width=\linewidth]{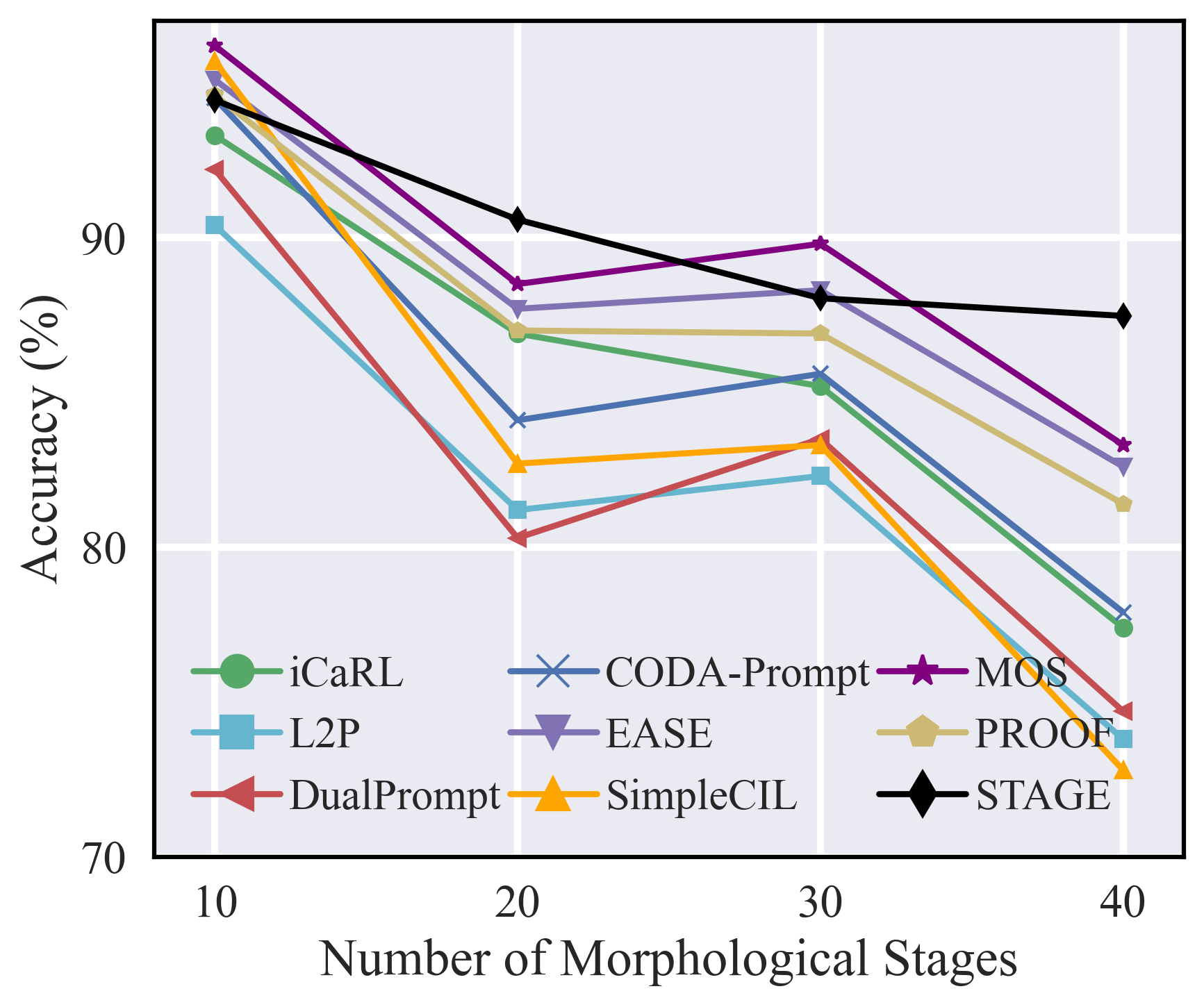}
    \caption{Insects}
  \end{subfigure}

  \caption{Performance curves across the ten Stage-Bench domains. All methods use the same pre-trained backbone and weights.}
  \label{fig:10domain}
\end{figure*}
\subsection{Per-Domain Performance on Stage-Bench}

To evaluate STAGE across diverse morphological evolution scenarios, we conduct a separate incremental-learning experiment on each Stage-Bench domain. Figure~\ref{fig:10domain} compares STAGE with eight methods from the main paper over the ordered two-stage stream of each domain.

Across the ten domains, STAGE maintains the highest or near-highest accuracy and generally shows a smaller drop after Stage-1 is introduced. The advantage is especially visible in domains with pronounced appearance changes. These results show that the aggregate improvement is not driven by only one or two domains and support explicit modeling of class-specific morphological change.

\paragraph{Drastic Evolution in the Insects Domain.}
The Insects domain includes pronounced larva-to-adult transformations and therefore directly probes the concern that anchor-based retrieval may favour appearances close to Stage-0. STAGE reaches $87.62\%$ final accuracy and $85.34\%$ Stage-1 accuracy on this domain, with $2.15\%$ Intra-F. Together with the residual interpretation above, these results indicate that the pool can represent large transformations without requiring the evolved appearance itself to resemble the initial anchor.

\subsection{Qualitative Visualization}

Figure~\ref{fig:supp_intra} reports Stage-0 and Stage-1 accuracy throughout the incremental stream. The narrower gap produced by STAGE indicates better preservation of the same class across morphological stages, consistent with the role of its evolution-aware memory pool. Figure~\ref{fig:supp_attention} provides a complementary Grad-CAM comparison. STAGE tends to emphasize stage-dependent regions, including body parts undergoing transformation or newly emerging structures, whereas PROOF more often focuses on the overall object. This qualitative evidence suggests that STAGE uses cues associated with morphological change rather than relying only on global object identity.

\section{Generalized Stage-CIL Protocol}

Stage-Bench instantiates Stage-CIL as a two-node linear evolution graph.
A more general formulation can associate each class $c$ with a directed
acyclic graph (DAG) $\mathcal{G}_c=(\mathcal{V}_c,\mathcal{E}_c)$, where
nodes $s \in \mathcal{V}_c$ represent morphological stages and edges
$(s \to s') \in \mathcal{E}_c$ encode feasible transitions. The current
benchmark uses $\mathcal{V}_c=\{0,1\}$ and
$\mathcal{E}_c=\{0\to 1\}$ for every class. A direct extension of STAGE
would apply its predict-then-classify operator along the observed edges
and maintain stage prototypes for the corresponding nodes. Our
three-stage Object-domain experiment evaluates the linear multi-stage
case; branching graphs remain a protocol and modeling direction rather
than an experimentally validated claim in this work.

\subsection{Multi-Stage Intra-Class Forgetting}
For $|\mathcal{S}_i|>2$, measuring only Stage-0 retention can miss forgetting at intermediate stages. We therefore additionally average the absolute before--after degradation over every non-terminal stage:
\begin{equation}
\begin{aligned}
F_{\mathrm{intra}}^{\mathrm{multi}}
&=\frac{1}{N_{\mathrm{cls}}}\sum_i
\frac{1}{|\mathcal{S}_i|-1}
\sum_{s=0}^{|\mathcal{S}_i|-2}
\left[A_{i,s}^{\mathrm{first}}-A_{i,s}^{\mathrm{final}}\right]_+ ,
\end{aligned}
\end{equation}
where $A_{i,s}^{\mathrm{first}}$ is the accuracy on stage $s$ immediately after it is learned and $A_{i,s}^{\mathrm{final}}$ is its accuracy at the end of the stream. On the three-stage Object-domain experiment, this multi-stage diagnostic is $0.35\%$ for STAGE and $2.15\%$ for MOS. It is distinct from the Stage-0-only Intra-F reported in the main table and confirms that the low forgetting score of STAGE is not obtained by retaining only Stage-0 while discarding the intermediate morphology.

\end{document}